\documentclass[acmsmall]{acmart}

\renewcommand\footnotetextcopyrightpermission[1]{} % removes footnote with conference information in first column
\usepackage{graphicx}
\usepackage{booktabs} % For formal tables
\usepackage{bbm}

\usepackage{array}
\usepackage{multirow}
\usepackage{url}
\usepackage{enumitem}
 
\usepackage{xspace}
 
\usepackage{cleveref}
\usepackage[normalem]{ulem}

\usepackage[utf8]{inputenc}
\usepackage[english]{babel}

\usepackage{scalerel}
\usepackage{stackengine,wasysym}
\usepackage{comment}

\newcommand\revision[1]{\color{black}{#1}\color{black}}

\usepackage{amsmath,amsthm,bm}
\usepackage{amsfonts}
\usepackage{algorithmic}
\usepackage[utf8]{inputenc}

\allowdisplaybreaks

\newcommand{\startpara}[1]{{\vskip1pt\noindent{\bf #1.}}}

\newcommand{\sectref}[1]{Section~\ref{#1}}
\newcommand{\figref}[1]{Figure~\ref{#1}}
\newcommand{\tabref}[1]{Table~\ref{#1}}

\newcommand{\defref}[1]{Definition~\ref{#1}}
\newcommand{\thref}[1]{Theorem~\ref{#1}}

\newtheorem{definition}{Definition}

\newtheorem{theorem}{Theorem}

\newcolumntype{L}[1]{>{\raggedright\let\newline\\\arraybackslash\hspace{0pt}}m{#1}}
\newcolumntype{C}[1]{>{\centering\let\newline\\\arraybackslash\hspace{0pt}}m{#1}}
\newcolumntype{R}[1]{>{\raggedleft\let\newline\\\arraybackslash\hspace{0pt}}m{#1}}

\newcommand{\always}{\square}
\newcommand{\eventually}{\lozenge}
\newcommand{\until}{\mathcal{U}_I}

\newcommand{\tablefontsize}{\footnotesize}

\newcommand{\traceset}{\omega}

\newcommand{\sat}{\models}
\newcommand{\satstrong}{\models_s}
\newcommand{\satweak}{\models_w}
\newcommand{\notsatstrong}{\not\models_s}
\newcommand{\notsatweak}{\not\models_w}
\newcommand{\notsat}{\not\models}
\newcommand{\eqdef}{\Leftrightarrow}

\newcommand{\conflevel}{\varepsilon}

\newcommand{\conflevelFun}{\epsilon}

\newcommand{\lb}{\Phi_t^-(\conflevel)}
\newcommand{\ub}{\Phi_t^+(\conflevel)}

\newcommand{\Rset}{\mathbb{R}}
\newcommand{\Tset}{\mathbb{T}}

\newcommand{\uncerF}{\mathcal{M}}

\usepackage[ruled]{algorithm2e}
\SetAlFnt{\small}
\SetAlCapFnt{\small}
\SetAlCapNameFnt{\small}
\SetAlCapHSkip{0pt}
\IncMargin{-\parindent}

\AtBeginDocument{%
  \providecommand\BibTeX{{%
    \normalfont B\kern-0.5em{\scshape i\kern-0.25em b}\kern-0.8em\TeX}}}

\acmBooktitle{}
\begin{document}

\title[Predictive Monitoring with Logic-Calibrated Uncertainty for CPS]{Predictive Monitoring with Logic-Calibrated Uncertainty for Cyber-Physical Systems}

 
 \author{Meiyi Ma}
 \affiliation{\institution{University of Virginia}
   \country{USA}}
 \email{{meiyi}@virginia.edu}
\author{John Stankovic}
 \affiliation{\institution{University of Virginia}
   \country{USA}}
 \email{stankovic@virginia.edu}

%  \author{Lu Feng}
%  \affiliation{\institution{University of Virginia}}
%  \email{lu.feng@virginia.edu}
 
  \author{Ezio Bartocci}
 \affiliation{\institution{TU Wien}
   \country{Austria}}
 \email{ezio.bartocci@tuwien.ac.at}

 \author{Lu Feng}
 \affiliation{\institution{University of Virginia}
   \country{USA}}
 \email{lu.feng@virginia.edu}

\renewcommand{\shortauthors}{Meiyi Ma, et al.}

\begin{abstract}
Predictive monitoring---making predictions about future states and monitoring if the predicted states satisfy requirements---offers a promising paradigm in supporting the decision making of Cyber-Physical Systems (CPS). Existing works of predictive monitoring mostly focus on monitoring individual predictions rather than sequential predictions. We develop a novel approach for monitoring sequential predictions generated from Bayesian Recurrent Neural Networks (RNNs) that can capture the inherent uncertainty in CPS, drawing on insights from our study of real-world CPS datasets. We propose a new logic named \emph{Signal Temporal Logic with Uncertainty} (STL-U) to monitor a flowpipe containing an infinite set of uncertain sequences predicted by Bayesian RNNs. We define STL-U strong and weak satisfaction semantics based on if all or some sequences contained in a flowpipe satisfy the requirement. We also develop methods to compute the range of confidence levels under which a flowpipe is guaranteed to strongly (weakly) satisfy an STL-U formula. Furthermore, we develop novel criteria that leverage STL-U monitoring results to calibrate the uncertainty estimation in Bayesian RNNs. Finally, we evaluate the proposed approach via experiments with real-world datasets and a simulated smart city case study, which show very encouraging results of STL-U based predictive monitoring approach outperforming baselines. 

\end{abstract}




\maketitle

\section{Introduction}
\label{sec:intro}

% what is predictive monitoring 
% why need predictive monitoring for CPS?
% state of the art and limitations

\emph{Predictive monitoring} concerns the problem of (continuously) making predictions about future states and monitoring if the predicted states satisfy or violate requirements.
Predictive monitoring offers a promising paradigm in supporting the decision making of Cyber-Physical Systems (CPS), for example, reducing an automated insulin delivery system's dosage if a potentially dangerous hypoglycemic condition is predicted, and adapting a traffic control system's signaling if traffic congestion due to car accidents or inclement weather is forecast.
On the one hand, various machine learning and statistical analysis techniques (e.g., neural networks, ARIMA) have been popularly applied to predict future states of CPS across different application domains, such as predicting glucose levels for artificial pancreas systems~\cite{montaser2017stochastic}, predicting takeover reaction time for automated vehicles~\cite{pakdamanian2020deeptake},
forecasting air quality~\cite{liang2018geoman}, fire risk~\cite{singh2018dynamic}, and frost damage~\cite{zhou2020frost} in smart cities.
On the other hand, there have been great efforts over the past decades devoted to develop runtime monitoring techniques and tools. 
For example, a survey of specification (e.g., Signal Temporal Logic (STL)~\cite{MalerN04}) based runtime monitoring of CPS is provided in~\cite{chapter5}.
Nevertheless, research on predictive monitoring that addresses challenges arisen from combining these two aspects has received scant attention until very recently. 
Existing works of predictive monitoring (e.g., \cite{bortolussi2019neural,babaee2019accelerated}) mostly focus on monitoring individual predictions rather than sequential predictions. 
A more recent work~\cite{qin2020clairvoyant} considers STL-based monitoring for predictions made from statistical time-series analysis, assuming that a joint probability distribution of predictions over multiple time-points can be estimated.

In this paper, we develop a novel approach for monitoring sequential predictions generated from Bayesian Recurrent Neural Networks (RNN) models.
RNN-based sequential prediction has been widely used in CPS applications (e.g., \cite{he2020towards, liu2019contextualized}).
Many commonly used RNN models (e.g., LSTM) are deterministic, which generate the same sequence of predictions given the same set of historical states.
A key challenge is how to generate and monitor predictions that can capture the inherent uncertainty in CPS (e.g., due to sensing noise, human interactions).
We study two real-world CPS datasets to analyze the uncertainty characteristics and implications on predictive monitoring. 
Insights from our study show that (i) deterministic RNN models are not suitable for representing the significant uncertainty exhibited in CPS, and (ii) there is a need for developing new monitors for checking sequential predictions with high uncertainty.

To address the first insight, we apply stochastic regularization techniques (SRTs)~\cite{gal2016dropout} to cast deterministic RNNs as Bayesian RNNs, which adapt deterministic sequential predictions as a sequence of posterior probability distributions to estimate the uncertainty. 
We formally define a \emph{flowpipe} signal to represent uncertain sequential predictions generated by Bayesian RNNs. 
The projection of a flowpipe for a single time-point is a confidence interval induced from 
a Gaussian distribution, which includes values of all possible sequences predicted by the Bayesian RNN.
A larger confidence interval indicates a higher level of uncertainty about the prediction. 

To address the second insight, we propose a new logic named \emph{Signal Temporal Logic with Uncertainty} (STL-U). Existing temporal logic based monitors (e.g., STL and its variants) mostly focus on deterministic signals and cannot be directly applied for monitoring an infinite set of sequences contained in a flowpipe. 
Several recent works (e.g., \cite{JhaRSS18,SadighK16,STSTL,STSTL2}) extend STL with stochastic predicates to reason about uncertainty. 
Our approach differs from these previous works fundamentally. 
Instead of reasoning about the probability of satisfying a predicate, 
STL-U checks a flowpipe signal containing an infinite set of sequences. 
For example, consider a STL-U formula $\always_{[0,2]}\mathsf{AQI}_{\conflevel=95\%}<50$, which represents 
the requirement ``the predicted Air Quality Index under 95\% confidence level should never exceed 50 in the next two hours''. We develop a STL-U monitor that checks if all (\emph{resp.} some) sequences contained in the predicated flowpipe satisfy the requirement, which we call STL-U strong (\emph{resp.} weak) satisfaction.
In addition, we equip the STL-U monitor with the capability to answer queries such as ``Under what confidence level, the predicated flowpipe is guaranteed to strongly (weakly) satisfy the STL-U formula?''
It is particularly useful to compute such confidence guarantees when users do not know \emph{a priori} about the level of prediction uncertainty. 

Furthermore, the quality of predictive monitoring results depends on the uncertainty estimated from Bayesian RNNs, which varies based on the choice of uncertainty estimation schemas (e.g., SRTs and dropout rates). 
In the current practice, an uncertainty estimation schema is often selected empirically or guided by traditional deep learning metrics (e.g., mean square error, negative log-likelihood, KL divergence), which tend to over-estimate the uncertainty level~\cite{gal2017concrete,xiao2019quantifying}.
In addition, these metrics treat the uncertainty estimation of each individual value in a predicted sequence separately, and thus lack an integrated view about the uncertainty of the sequence. 
To address this limitation, we develop novel criteria that leverage STL-U monitoring results to select and tune uncertainty estimation schemas. Such STL-U criteria can help to calibrate the uncertainty estimates for predictive monitoring. 

We compare STL-U criteria with state-of-the-art baselines via experimental evaluation on real-world CPS datasets. The results are very promising: STL-U criteria outperform all six baselines in terms of F1-scores comparing STL-U monitoring results for the predicted and target sequences. 
In addition, experiments also show that STL-U uncertainty calibration is compatible with different types of RNN models.

Finally, we evaluate the STL-U based predictive monitoring approach via a simulated smart city case study with 10 smart services and 390 requirements. 
Experiment results demonstrate the efficiency of the approach. 
In one case, it only takes around 281 seconds to monitor 130,000 predicted flowpipes. 
Moreover, our approach can better support decision making in the simulated smart city. The simulation results show that our approach improves various city performance metrics (e.g., emergency waiting time, vehicle waiting number) significantly when compared with two baselines. 

\startpara{Contributions}
We summarize the major contributions of this paper as follows.

\begin{itemize}
    \item We develop a novel STL-U based predictive monitoring approach for CPS, which continuously monitors uncertain sequential predictions about future states generated by Bayesian RNN models.
    \item We create novel STL-U criteria for calibrating uncertainty estimation in Bayesian deep learning. 
    \item We evaluate the proposed approach via real-world smart city datasets and a simulated smart city case study, which show encouraging results. 
\end{itemize}

\startpara{Paper Organization}
In the rest of the paper, 
we describe the motivating study of CPS uncertainty in \sectref{sec:motivation},
provide an overview of STL-U based predictive monitoring approach in \sectref{sec:overview},
propose STL-U logic and monitoring algorithms in \sectref{sec:monitor},
present STL-U criteria for uncertainty calibration in \sectref{sec:prediction},
describe the evaluation results in \sectref{sec:evaluation},
discuss the related work in \sectref{sec:related},
and draw conclusions in \sectref{sec:conclusion}.

\section{Motivating Study}
\label{sec:motivation}

In this section, we study the following real-world smart city datasets as motivating examples
to analyze uncertainty characteristics and to discuss implications on predictive monitoring for CPS. 

\begin{enumerate}
    \item \emph{Air quality dataset}~\cite{liang2018geoman} collected by Microsoft Research from 437 air quality monitoring stations in China during the period of 5/1/2014 to 4/30/2015, which includes  2,891,393 records of air quality index (AQI).
    \item \emph{Traffic volume dataset}~\cite{nycopendata} collected by the NYC Department of Transportation from 1,490 street segments in the New York City during the period of 9/13/2014 to 4/5/2018, which includes 514,776 records of traffic volume count. 
\end{enumerate}

\begin{figure}[t]
\includegraphics[width=13cm]{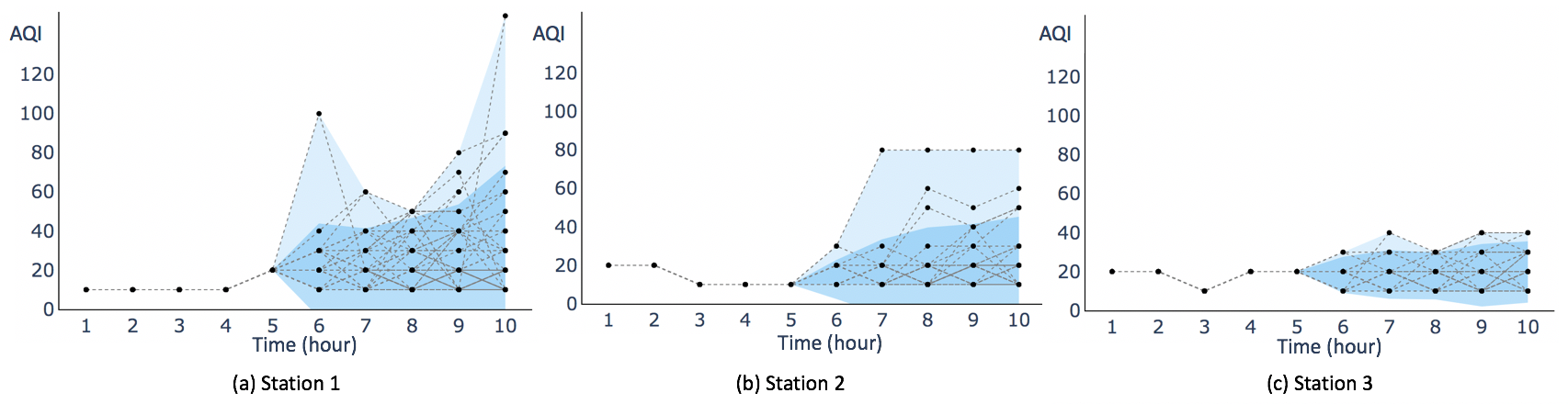}
\vspace{-0.2cm}
\caption{The uncertainty level varies across different stations in the air quality dataset.} 
\label{fig:bjuncertainty}
\vspace{-0.1cm}
\end{figure}

\begin{figure}[t]
\includegraphics[width=13cm]{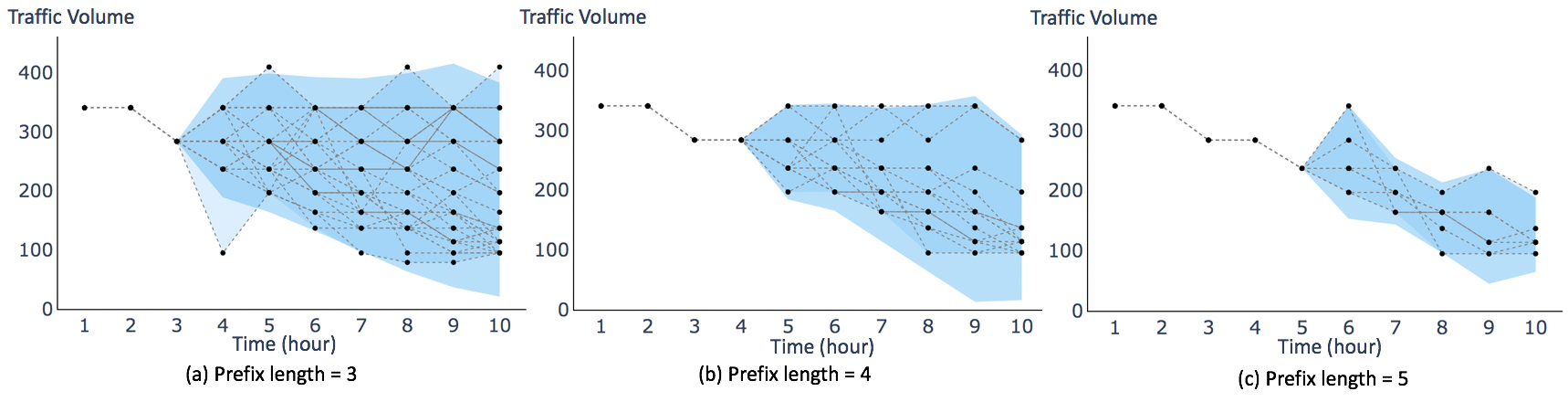}
\vspace{-0.2cm}
\caption{The uncertainty level varies for different pre-knowledge (prefix lengths) in the traffic volume dataset. } 
\label{fig:nycuncertainty}
% \vspace{-0.5cm}
% (the dark blue shadow covers the range of 95\% percentile.)
\end{figure}

\startpara{Uncertainty characteristics}
We made the following observations by analyzing these datasets.
\begin{itemize}
    \item \emph{Significant uncertainty exists in smart cities and the uncertainty level varies across different locations.} As an illustrative example, \figref{fig:bjuncertainty} shows 10-hour data segments taken from three different stations in the air quality dataset. 
    We preprocessed the raw data by averaging the data within an hour and performing a uniform quantization~\cite{kay1993fundamentals}. %where $Q(x) = \left \lceil{x/10}\right \rceil$.
    \figref{fig:bjuncertainty} plots data segments with the same prefixes (i.e., the same average AQI levels) for the first five hours.
    However, these data segments show significant uncertainty in the suffixes. 
    The light shadows in the figure cover the entire data range, and the dark shadows represent the range of 95\% percentile~\footnote{We utilize 95\% percentile because it is commonly used to represent the majority of the population distributed~\cite{sachs2012applied}.} of the corresponding normal distribution at a time. A larger range of 95\% percentile indicates a higher level of data uncertainty. 
    Thus, \figref{fig:bjuncertainty} shows that station 1 has the highest uncertainty level, followed by station 2 and station 3. 
    \item \emph{The data uncertainty level is affected by the pre-knowledge (i.e., the prefix length of data segments).} As an illustrative example, \figref{fig:nycuncertainty} plots 10-hour data segments taken from the same location in the traffic volume dataset. We preprocessed the data by 
    averaging the traffic volume counts within an hour and performing a logarithmic quantization~\cite{kay1993fundamentals}. %$Q(x) = \left \lceil{\log_{1.2} (x)}\right \rceil$
    \figref{fig:nycuncertainty} shows that, as the length of common data segment prefixes increases (i.e., more pre-knowledge about the data), the uncertainty level reduces. 
\end{itemize}

\revision{The uncertainty in CPS data could arise from many sources, such as noise from the sensing data (e.g., reading errors, faults, anomalies), the environment (e.g., unexpected weather or events like accidents), and human behaviors (e.g., interventions from human operators), to name a few. Thus, predictive monitoring for CPS should account for the impact of uncertainty.} 

%----------------------------------
\begin{table}[t]
\caption{Number of satisfying data segments for each STL formula.}
\vspace{-0.1cm}
\label{tab:datareq}
% \tablefontsize
\small
\centering
\begin{tabular}{l|ccccccccc}
\toprule
STL formulas & $\lambda=50$   & $\lambda=51$   & $\lambda=70$   & $\lambda=75$   & $\lambda=80$   \\ \hline 
$\always_{[0,10]}\mathsf{AQI}<\lambda$ & 2,807 & 2,895 & 3,614 & 4,011 & 4,443  \\ 
$\eventually_{[0,10]}\mathsf{AQI}<\lambda$ & 6,670 & 6,731 & 7,304 & 7,408 & 7,493  \\ 
$\mathsf{AQI}<150 \ \mathcal{U}_{[0,10]} $  $\mathsf{AQI}<\lambda$ & 5,558 & 5,613 & 6,030 & 6,169 & 6,230  \\
% \revision{$\eventually_{[0,5]}\always_{[0,5]}\mathsf{AQI}<\lambda$} & &&&&\\
$\always_{[0,10]}\mathsf{Traffic}<\lambda$ & 1,241 & 1,359 & 3,090 & 3,332 & 3,532  \\ 
$\eventually_{[0,10]}\mathsf{Traffic}<\lambda$ & 4,220 & 4,283 & 4,546 & 4,563 & 4,598 \\ 
% \revision{$\always_{[0,5]}\eventually_{[0,5]}\mathsf{Traffic}<\lambda$} & &&&&\\
\bottomrule
\end{tabular}
% \vspace{-0.3cm}
\end{table}
%----------------------------------

\startpara{Implications on predictive monitoring}
We discuss how the uncertainty in CPS would affect predictive monitoring from two aspects: (i) prediction, and (ii) monitoring. 

First, existing deterministic prediction models (e.g., RNNs) mostly forecast future states based on historical states. Given the same historical data, a deterministic model always yields the same prediction about future states. 
However, as discussed above, real-world CPS data exhibits significant uncertainty. For example, \figref{fig:bjuncertainty} illustrates that data segments with the same average AQI levels for the first five hours can lead to a diverse range of trends for the following five hours. 
Thus, deterministic prediction models are not suitable to capture the uncertainty in CPS data. 
There is a need for developing new techniques that can predict future states with appropriate levels of uncertainty.

Second, existing works (e.g., \cite{ma2018cityresolver}) have applied Signal Temporal Logic (STL) to specify and monitor city requirements. 
For example, a requirement that ``the AQI level within 10 hours should always be below certain threshold $\lambda$'' can be specified with a STL formula $\always_{[0,10]}(\mathsf{AQI}<\lambda)$,
where $\always$ is the temporal logical operator representing ``always'' and $\lambda$ is a parameter (e.g., $\lambda=50$ for good air quality).
\tabref{tab:datareq} shows five example STL formulas, with the first three representing city requirements about AQI and the last two representing city requirements about traffic volume. 
We applied a STL monitor to check how many 10-hour data segments of a selected location in the air quality and traffic volume datasets satisfy these STL formulas with varying parameter values of $\lambda$. 
We observe from \tabref{tab:datareq} that the STL monitoring results can be very sensitive to the change of $\lambda$ values. 
For example, the number of satisfying data segments for $\eventually_{[0,10]}(\mathsf{AQI}<\lambda)$ increases by 61 (from 6,670 to 6,731) when the $\lambda$ value only increases by 1 (from 50 to 51), and goes up 85 (from 7,408 to 7,493) when the $\lambda$ value increases from 75 to 80. 
Even though the differences also vary by the type of requirements, the amount is still too large to ignore. From the perspective of the data, it shows that a small difference of the data could completely change the monitoring results. However, it is impossible to predict the data with 100\% accuracy due to the existence of uncertainty in CPS, which makes the monitoring results less effective to support decision making. 
It also indicates that the existing monitors are not suitable for data with high uncertainty. Therefore, there is a need for developing new monitors that can check the prediction results accounting for the uncertainty.

\begin{figure}[t]
    \centering
   \includegraphics[width=\textwidth]{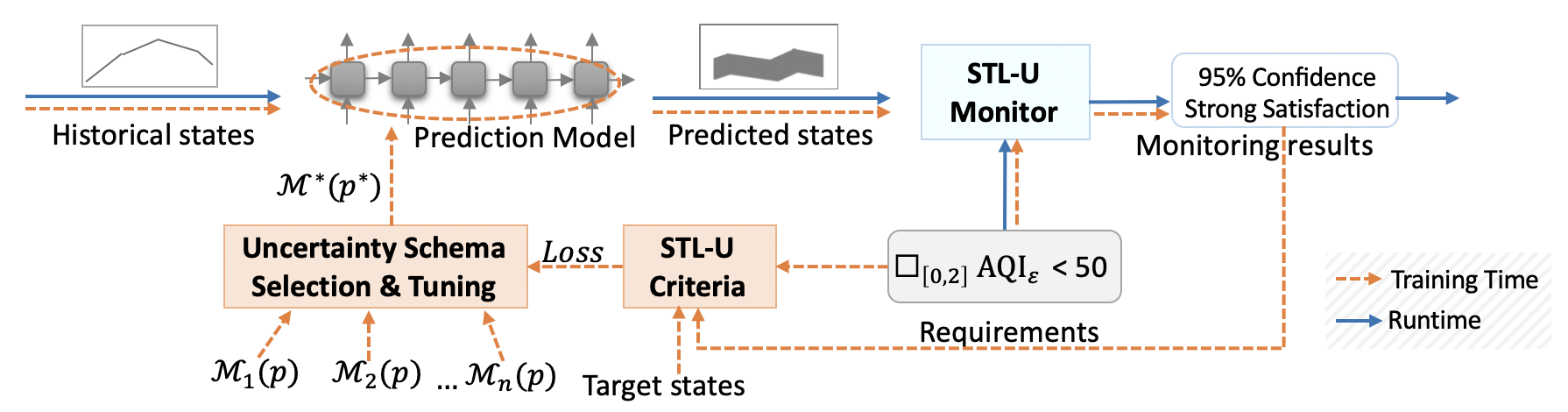}
    % ,height=0.56\textwidth
    % \vspace{-0.5cm}
    \captionof{figure}{Overview of STL-U based predictive monitoring approach.}
    \label{fig:overview}
    \vspace{-0.4cm}
\end{figure}

% \begin{figure}[t]
%     \centering
%     \includegraphics[width=12.5cm]{Figure_EmSoft/AQI_EX.png}
%     \caption{\blue{An Example of AQI Prediction (delete?)}}
%     \label{fig:AQI_example}
%     % \vspace{-0.5cm}
% \end{figure}

\section{Approach Overview}
\label{sec:overview}
We develop a novel predictive monitoring approach for CPS, as illustrated in \figref{fig:overview}, to address limitations discussed in the previous section.
Our approach adopts Bayesian RNN models to predict future states (e.g., AQI in next 2 hours) based on historical data (e.g., AQI in the past 5 hours). 
By contrast to deterministic prediction models that output a single sequence of predicted values, Bayesian RNN models generate a sequence of distributions to capture the uncertainty of predicted future states, which are represented by a range of potential values under a given confidence level at each time-point. 
We propose a new logic named  \emph{Signal Temporal Logic with Uncertainty} (STL-U) and develop a STL-U monitor to check such uncertain sequential predictions generated by Bayesian RNN models.
STL-U is expressive enough to specify CPS requirements with uncertainty confidence levels.
For example, a STL-U formula $\always_{[0,2]}\mathsf{AQI}_{\conflevel=95\%}<50$ represents the requirement ``the predicted AQI under 95\% confidence level should never exceed 50 in the next two hours''.
The STL-U monitor checks if all or some possible sequences of future states predicted by the Bayesian RNN model satisfy the requirement, which we call strong and weak satisfaction relations. 
When the confidence level is unspecified in a formula (e.g., $\always_{[0,2]}\mathsf{AQI}_{\conflevel=?}<50$), we can also use STL-U monitor to compute the range of confidence levels under which the predicted flowpipe is guaranteed to strongly or weakly satisfy a requirement. 
In addition, we develop novel criteria (loss functions) based on STL-U monitoring results to calibrate the uncertainty estimation in Bayesian deep learning.

At \textit{training} time (the flow marked by orange dash-lines in \figref{fig:overview}),
the proposed approach automatically selects and tunes an optimal uncertainty estimation schema based on STL-U criteria.
As we will discuss in \sectref{sec:prediction}, such uncertainty calibration is an essential step to guarantee the quality of predictive monitoring, in order to better support decision making of CPS. 
At \textit{runtime} (the flow marked by the blue lines in \figref{fig:overview}), the proposed approach runs as a continuous iterative process to monitor the predicted future states. 
Considering the predictive monitoring of AQI in a smart city, for example, 
at time $t$, the proposed approach first predicts the AQI for the future 3 hours from time $t$ and monitors if the predictions satisfy the requirements; after a period $\Delta t$ (e.g., 30 minutes), it predicts the AQI for the future 3 hours from $t+\Delta t$ and checks if the new predictions satisfy the requirements. 
In this way, the proposed approach provides continuous predictive monitoring of future states to support decision making of CPS.

\section{STL-U Monitor}
\label{sec:monitor}

As described in the previous section, our approach adopts Bayesian RNN models to make sequential predictions about uncertain future states. 
We propose a \emph{Signal Temporal Logic with Uncertainty} (STL-U) to monitor such uncertain sequential predictions. 
We introduce STL-U syntax and semantics in \sectref{sec:stl-u}, and present methods to compute STL-U confidence guarantees in \sectref{sec:confidence}.

%=================================================================================
\subsection{STL-U Syntax and Semantics}\label{sec:stl-u}
We formally define a new type of signals called \emph{flowpipes}\footnote{To be noted, here we use the concept of flowpipes but define it in a new way.} to represent uncertain sequential predictions generated by Bayesian RNN models. We describe more details about Bayesian RNN models and uncertainty estimation later in \sectref{sec:prediction}.

\begin{definition}[Flowpipe]\label{def:flowpipe}
A single-variable flowpipe $\Omega$ is defined over a finite discrete time domain $\Tset$ such that $\Omega[t] = \Phi_t$ at any time $t\in\Tset$ and $\Phi_t$ is a Gaussian distribution $\mathcal{N}(\theta_t, \sigma^2_t)$.
Let $\omega: \{\Omega\}^n$ be a (multi-variable) flowpipe signal,
where $n=|X|$ is the size of a finite set of (independent) real variables $X$.
Each variable $\mathsf{x} \in X$ has a corresponding flowpipe $\omega_\mathsf{x}$ whose value at time $t$ follows a Gaussian distribution $\Phi_t$, denoted by $\omega_\mathsf{x}[t]=\Phi_t$.
\end{definition}

% \noindent
% Note that in the above definition, the mean $\theta_t$ and the variance $\sigma_t$ of Gaussian distribution can be computed following equations  in Section~\ref{sec:uncertainty}.

Given a confidence level $\conflevel \in (0,1) \subseteq \Rset$, a single-variable flowpipe $\Omega$ at time $t$ is bounded by a confidence interval $[\lb, \ub]$ with 
the lower bound $\lb = \theta_t - \delta \cdot \frac{\sigma_t}{\sqrt{N}}$ 
and the upper bound $\ub = \theta_t + \delta \cdot \frac{\sigma_t}{\sqrt{N}}$,
where \revision{$N$ is the number of samples that the Gaussian distribution is estimated from, and 
$\delta$ is a function ${\delta=F^{-1}(\frac{\conflevel}{2})}$ with $F$ denoting the CDF of the standard normal distribution $\mathcal{N}(0,1)$~\cite{sachs2012applied}. } 
% is a parameter obtained from the quantile function of Gaussian distribution $\Phi_t$ based on the confidence level $\conflevel$. 
In the special case where the Gaussian distribution's variance is $\sigma_t=0$,
a flowpipe signal becomes a single trace because the lower and upper bounds of the confidence interval coincide (i.e., $\lb=\ub=\theta_t$). 
Given a (multi-variable) trace $\bar{\omega}$ and a flowpipe $\omega$ over the same set of real variables $X$, we say that $\bar{\omega}$ belongs to $\omega$, denoted by $\bar{\omega} \in \omega$, if $\bar{\omega}_\mathsf{x}[t] \in [\lb, \ub]$ for all $\mathsf{x} \in X$ and $t\in \Tset$, 
where $[\lb, \ub]$ is the confidence interval of flowpipe $\omega_\mathsf{x}$ under confidence level $\conflevel$.

\begin{definition}[STL-U Syntax]
A STL-U formula $\varphi$ over a flowpipe signal $\omega$ is given by
$$
\varphi := \mu_\mathsf{x}(\conflevel) \ |\ \neg \varphi \ |\ \varphi_1 \land \varphi_2 \ |\ \always_I \varphi\ |\ \eventually_I \varphi \ |\ \varphi_1 \ \until \ \varphi_2
$$
where $\mu_\mathsf{x}(\conflevel)$ is an atomic predicate over variable $\mathsf{x}$ with confidence level $\conflevel$, whose value is determined by $\mu_\mathsf{x}(\conflevel) \equiv f(x)>0$ with a continuous function $f(x)$ about  flowpipe $\omega_\mathsf{x}$ under confidence level $\conflevel$.
Temporal operators $\always_I$, $\eventually_I$ and $\until$ with a time interval $I\subseteq \Tset$ represent (bounded) ``always'', ``eventually'', and ``until'', respectively. 
\label{def:syntax}
\end{definition}

%=================================================================================

%-----------------------------------------
\noindent
\begin{minipage}{0.611\textwidth}
\centering
% \begin{figure}
% \centering
\includegraphics[width=\textwidth]{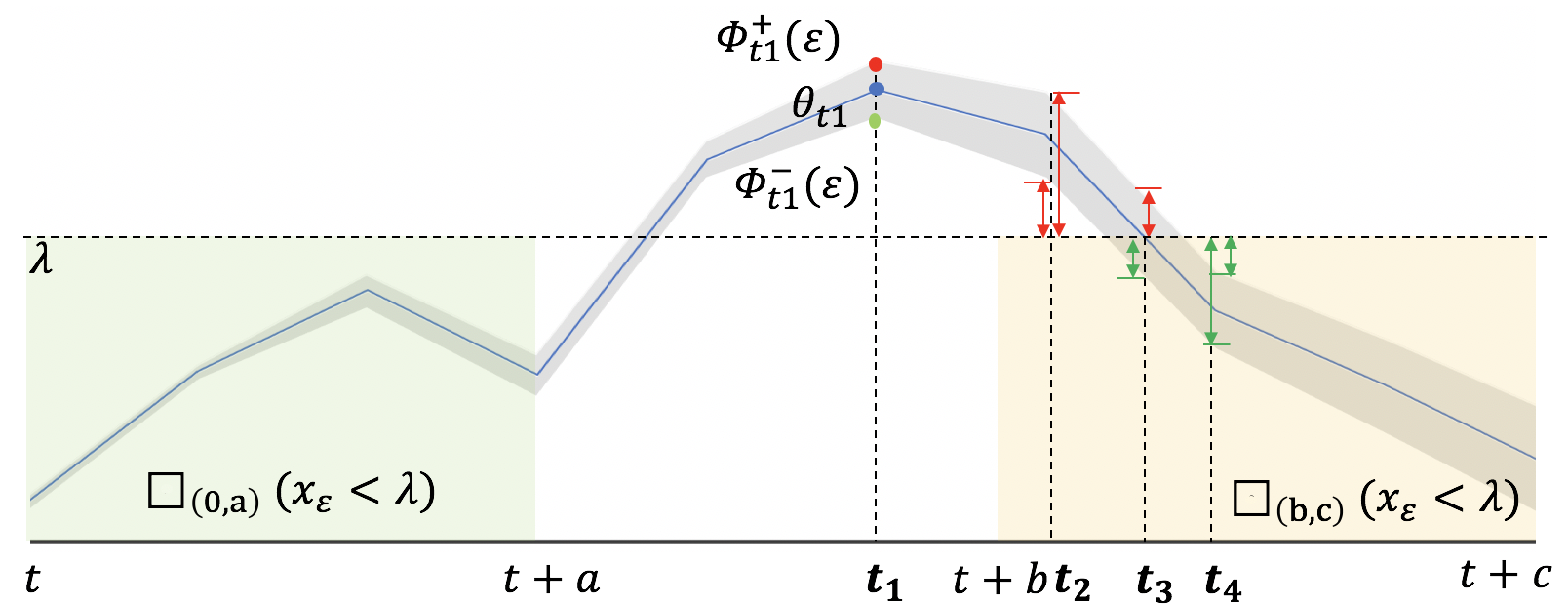}
\captionof{figure}{An example flowpipe under the confidence level $\conflevel$.} 
\label{fig:demo}
% \vspace{-0.5cm}
% \end{figure}
\end{minipage}
% \vspace{0.5cm}
\begin{minipage}{0.389\textwidth}
% \begin{figure}
    \centering
    % \vspace{0.1cm}
    \includegraphics[width=\textwidth]{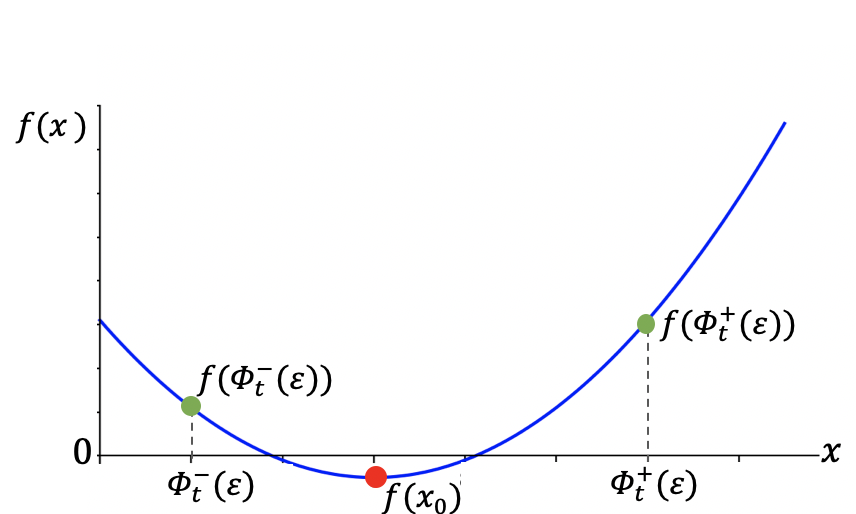}
    \captionof{figure}{An example function $f(x)$.}
    \label{fig:f_opt}
% \end{figure}
\end{minipage}
\vspace{0.5cm}

%-----------------------------------------

%\startpara{STL semantics}
We define the semantics of a flowpipe signal $\omega$ satisfying a STL-U formula $\varphi$ at time $t$ by two indices:
\emph{strong satisfaction}, denoted by $(\traceset, t) \satstrong \varphi$;  
and \emph{weak satisfaction}, denoted by $(\traceset, t) \satweak \varphi$.

\begin{definition} STL-U strong satisfaction semantics.
{ 
\begin{alignat*}{2}
    (\traceset, t) & \satstrong \mu_\mathsf{x}(\conflevel) 
        && \eqdef  \forall x \in [\lb, \ub], f(x)>0 \\    
    (\traceset, t) & \satstrong \neg \varphi 
        && \eqdef   (\traceset, t) \notsatweak \varphi \\
    (\traceset, t) & \satstrong \varphi_1 \land \varphi_2 
        && \eqdef  (\traceset, t) \satstrong \varphi_1 \mbox{ and } (\traceset, t) \satstrong \varphi_2  \\
        (\traceset, t) & \satstrong \always_I \varphi 
        && \eqdef  \forall t' \in (t+I), (\traceset, t') \satstrong \varphi \\
            (\traceset, t) & \satstrong \eventually_I \varphi 
        && \eqdef   \exists t' \in (t+I), (\traceset, t') \satstrong \varphi \\
    (\traceset, t) & \satstrong \varphi_1 \until \varphi_2 
        && \eqdef \exists t' \in (t+I) \cap \Tset, (\traceset, t') \satstrong \varphi_2    \mbox{ and }  
         \forall t'' \in (t, t'), (\traceset, t'') \satstrong \varphi_1\\
\end{alignat*}
}
\label{def:strongboolean}
\vspace{-1cm}
\end{definition}

\begin{definition} STL-U weak satisfaction semantics.
{
\begin{alignat*}{2}
    (\traceset, t) & \satweak \mu_\mathsf{x}(\conflevel) 
        && \eqdef \exists x \in [\lb, \ub], f(x)>0\\
    (\traceset, t) & \satweak \neg \varphi 
        && \eqdef  (\traceset, t) \notsatstrong \varphi  \\
    (\traceset, t) & \satweak \varphi_1 \land \varphi_2 
        && \eqdef  (\traceset, t) \satweak \varphi_1 \mbox{ and } (\traceset, t) \satweak \varphi_2 \\
        % (\traceset, t) & \satweak \varphi_1 \land \varphi_2 
        % && \eqdef  \exists \traceset' \in \traceset, (\traceset, t) \satweak \varphi_1 \mbox{ and } (\traceset, t) \satweak \varphi_2  \\
            (\traceset, t) & \satweak \always_I \varphi 
        && \eqdef  \forall t' \in (t+I), (\traceset, t') \satweak \varphi \\
            (\traceset, t) & \satweak \eventually_I \varphi 
        && \eqdef   \exists t' \in (t+I), (\traceset, t') \satweak \varphi \\
    (\traceset, t) & \satweak \varphi_1 \until \varphi_2 
        && \eqdef \exists t' \in (t+I) \cap \Tset, (\traceset, t') \satweak \varphi_2   \mbox{ and } 
     \forall t'' \in (t, t'), (\traceset, t'') \satweak \varphi_1 \\
\end{alignat*}
}
\label{def:weakboolean}
 \vspace{-1.cm}
\end{definition}

\noindent
To be noted, the negation of strong satisfaction is equivalent to weak violation, and the negation of weak satisfaction is equivalent to strong violation. 
Intuitively, strong satisfaction means that all values bounded within the confidence interval of a flowpipe should satisfy the STL-U formula, while weak satisfaction means that there exist some value within the confidence interval of a flowpipe satisfying the STL-U formula. 
In CPS applications, strong satisfaction relations can be used for monitoring strict requirements (e.g., safety), while weak satisfaction relations can be used for monitoring soft constraints (e.g., energy consumption).

Figure~\ref{fig:demo} shows an example flowpipe signal under the confidence level $\conflevel$, representing predictions from time $t$ to $t+c$.  
At a time-point $t_1$, the flowpipe follows a Gaussian distribution $\Phi_{t_1}$ with the mean of $\theta_{t_1}$ and is bounded by a confidence interval $[\Phi_{t_1}^-(\conflevel), \Phi_{t_1}^+(\conflevel)]$.
This flowpipe signal strongly satisfies STL-U formula $\always_{(0,a)}(x_\conflevel<\lambda)$ at time $t$,  
because the flowpipe signal values bounded within the confidence interval from time $t$ to $t+a$ are all below the threshold $\lambda$ (see the left green zone in Figure~\ref{fig:demo}).
Consider another STL-U formula $\always_{(b,c)}(x_\conflevel<\lambda)$.
As shown in Figure~\ref{fig:demo} (yellow zone in the right), the flowpipe's confidence interval is entirely above the threshold $\lambda$ at time $t_2$, 
partially below $\lambda$ at time $t_3$, and entirely below $\lambda$ at time $t_4$.
Therefore, the flowpipe neither strongly nor weakly satisfies the STL-U formula $\always_{(b,c)}(x_\conflevel<\lambda)$ at time $t$.

\revision{
\begin{theorem}[Strength relation theorem]\label{th:correct}
If a flowpipe $\traceset$ strongly satisfies a STL-U formula $\varphi$ at time $t$, then the weak satisfaction relation also holds. On the other hand, if the flowpipe $\traceset$ does not weakly satisfy a STL-U formula $\varphi$ at time $t$, then it would also not strongly satisfy $\varphi$.
Formally, 
\begin{equation*}
\begin{array}{ll}
(\traceset, t) \satstrong \varphi &\Rightarrow  (\traceset, t) \satweak \varphi \\
(\traceset, t) \notsatweak \varphi & \Rightarrow  (\traceset, t) \notsatstrong \varphi \\
\end{array}
\end{equation*}
\end{theorem}

We include the proof of Theorem \ref{th:correct} and properties of STL-U semantics in the Appendix.
}

It is challenging to monitor STL-U strong and weak satisfactions for a flowpipe that contains an infinite set of sequences.
Take the atomic predicate $\mu_\mathsf{x}(\conflevel)$ as an example. 
Based on \defref{def:strongboolean}, a flowpipe strongly satisfies $\mu_\mathsf{x}(\conflevel)$ iff $f(x)>0$ for all $x \in [\lb, \ub]$.
It is computationally expensive if not infeasible to exhaustively search through the entire confidence interval.
In addition, it does not suffice to only check the lower and upper bounds of the confidence interval when $f(x)$ is a non-monotonic function. 
\figref{fig:f_opt} shows an example where $f(\lb)>0$ and $f(\ub)>0$, but there is a $x_0 \in [\lb, \ub]$ with $f(x_0)<0$. 
We tackle this challenge by computing the minimal value $f_\mathsf{min}$ of $f(x)$ for $x \in [\lb, \ub]$ (e.g., via minimization algorithms in~\cite{brent2013algorithms}).
If $f_\mathsf{min}>0$, which implies that $f(x)>0$ for all $x \in [\lb, \ub]$,
then the flowpipe strongly satisfies $\mu_\mathsf{x}(\conflevel)$.
Based on \defref{def:weakboolean}, a flowpipe weakly satisfies $\mu_\mathsf{x}(\conflevel)$ iff there exist some $x \in [\lb, \ub]$ such that $f(x)>0$.
For monitoring weak satisfaction, we compute the maximal value $f_\mathsf{max}$ of $f(x)$ for any $x \in [\lb, \ub]$ and check if $f_\mathsf{max}>0$.
We include pseudo code of monitoring algorithms for STL-U strong and weak satisfactions as Algorithm 1 and Algorithm 2 in the Appendix.
\revision{Following \defref{def:flowpipe}, we use an array of triplets $\langle t, \theta_t, \sigma_t  \rangle$, $t \in \Tset$ to represent a flowpipe in STL-U monitoring algorithms. Given $\theta$, $\sigma$, and $\conflevel$, calculating $\ub$ and $\lb$ takes a constant time $O(1)$, which could be further accelerated by caching the intermediate results. The time complexity of calculating the predicate $f(x)>0$ depends on the complexity of $f(x)$ and the selected minimization algorithms.  
The time complexity of STL-U monitoring algorithms is similar to STL monitoring algorithms.
Thus, STL-U can be used to monitor complex specifications (e.g., with multiple levels of nesting temporal operators) via using Algorithm 1 or Algorithm 2 recursively}.

%=================================================================================
\subsection{STL-U Confidence Guarantees}\label{sec:confidence}
It may not always be possible for users to specify a confidence level for a flowpipe \emph{a priori}.
It is therefore useful to query about, under what confidence level, a flowpipe is guaranteed to strongly (weakly) satisfy a STL-U formula. 
We present methods to compute such confidence guarantees as follows. 

Let $\conflevelFun_s(\varphi, \traceset, t)$ and $\conflevelFun_w(\varphi, \traceset, t)$
denote the range of confidence levels that guarantee a flowpipe signal $\traceset$ 
strongly and weakly satisfying a STL-U formula $\varphi$ at time $t$, respectively. 
Let $\conflevelFun_s^c$ (\emph{resp.} $\conflevelFun_w^c$) denotes the complement set of $\conflevelFun_s$ (\emph{resp.} $\conflevelFun_w$) within the interval $(0,1)$.

\begin{definition}
Confidence guarantees for STL-U strong satisfaction.
{
\begin{alignat*}{2}
    &\conflevelFun_s(\mu_\mathsf{x}, \traceset, t) && =  
\left (0, \int_{\theta_t - \eta}^{\theta_t + \eta} \Phi_t(x) dx\right ), \mathsf{where}\ \eta = \inf\left\{|x - \theta_t| \mathrel{\Big|} f(x) \leq 0 \right\}
% \emptyset, \text{if } \eta = 0
   \\
    &\conflevelFun_s(\neg \varphi, \traceset, t) && = \conflevelFun_w^c(\varphi, \traceset, t) \\
    &\conflevelFun_s(\varphi_1 \land \varphi_2, \traceset, t) && =  
    \conflevelFun_s(\varphi_1, \traceset, t) \cap \conflevelFun_s(\varphi_2, \traceset, t)   \\
 %   &\conflevelFun_s(\always_I \varphi, \traceset, t) && = \underset{t' \in (t, t+I)}{\cap} 
 %  \conflevelFun_s(\varphi, \traceset, t')\\
     &\conflevelFun_s(\always_I \varphi, \traceset, t) && = \underset{t' \in (t+I)}{\bigcap} 
  \conflevelFun_s(\varphi, \traceset, t')\\
      &\conflevelFun_s(\eventually_I \varphi, \traceset, t) && = \underset{t' \in (t+I)}{\bigcup} 
  \conflevelFun_s(\varphi, \traceset, t')\\
    &\conflevelFun_s( \varphi_1 \until \varphi_2, \traceset, t) && =
    \underset{t' \in (t+I)}{\bigcup}
    \left\{\conflevelFun_s(\varphi_2, \traceset, t')\cap(
    \underset{t'' \in (t, t')}{\bigcap} 
   \conflevelFun_s(\varphi_1, \traceset, t''))\right\} 
         \\
 \end{alignat*}
 }
\label{def:strongcl}
\vspace{-1.2cm}
\end{definition}

\begin{definition}
Confidence guarantees for STL-U weak satisfaction.
{
\begin{alignat*}{2}
&\conflevelFun_w(\mu_\mathsf{x}, \traceset, t) && = \left( \int_{\theta_t - \eta}^{\theta_t + \eta} \Phi_t(x) dx, 1 \right ),~\mathsf{where}\ \eta = \inf\left\{|x - \theta_t| \mathrel{\Big|} f(x) > 0 \right\} \\
    &\conflevelFun_w(\neg \varphi, \traceset, t) && = \conflevelFun_s^c(\varphi, \traceset, t) \\
    &\conflevelFun_w(\varphi_1 \land \varphi_2, \traceset, t) && =  
    \conflevelFun_w(\varphi_1, \traceset, t) \cap \conflevelFun_w(\varphi_2, \traceset, t)   \\
    &\conflevelFun_w(\always_I \varphi, \traceset, t) && = \underset{t' \in (t+I)}{\bigcap} 
  \conflevelFun_w(\varphi, \traceset, t')\\
      &\conflevelFun_w(\eventually_I \varphi, \traceset, t) && = \underset{t' \in (t+I)}{\bigcup} 
  \conflevelFun_w(\varphi, \traceset, t')\\
    &\conflevelFun_w( \varphi_1 \until \varphi_2, \traceset, t) && =
    \underset{t' \in (t+I)}{\bigcup}
    \left\{\conflevelFun_w(\varphi_2, \traceset, t') \cap ( 
    \underset{t'' \in (t, t')}{\bigcap} 
   \conflevelFun_w(\varphi_1, \traceset, t''))\right\}
         \\
\end{alignat*}
}
\label{def:weakcl}
\vspace{-1cm}
\end{definition}
% \vspace{-1cm}

\begin{theorem}\label{th:strong}
Given a flowpipe signal $\omega$ and a STL-U formula $\varphi$, $\omega$ is guaranteed to strongly satisfy $\varphi$ at time $t$ under a confidence level $\conflevel \in \conflevelFun_s(\varphi, \traceset, t)$ computed based on \defref{def:strongcl}.
\end{theorem}

\begin{theorem}\label{th:weak}
Given a flowpipe signal $\omega$ and a STL-U formula $\varphi$, $\omega$ is guaranteed to weakly satisfy $\varphi$ at time $t$ under a confidence level $\conflevel \in \conflevelFun_w(\varphi, \traceset, t)$ computed based on \defref{def:weakcl}.
\end{theorem}

% \noindent 
We include proofs of the above theorems in the Appendix.
% \vspace{0.2cm}

\begin{figure}[b]
\centering
\includegraphics[width=0.95\textwidth]{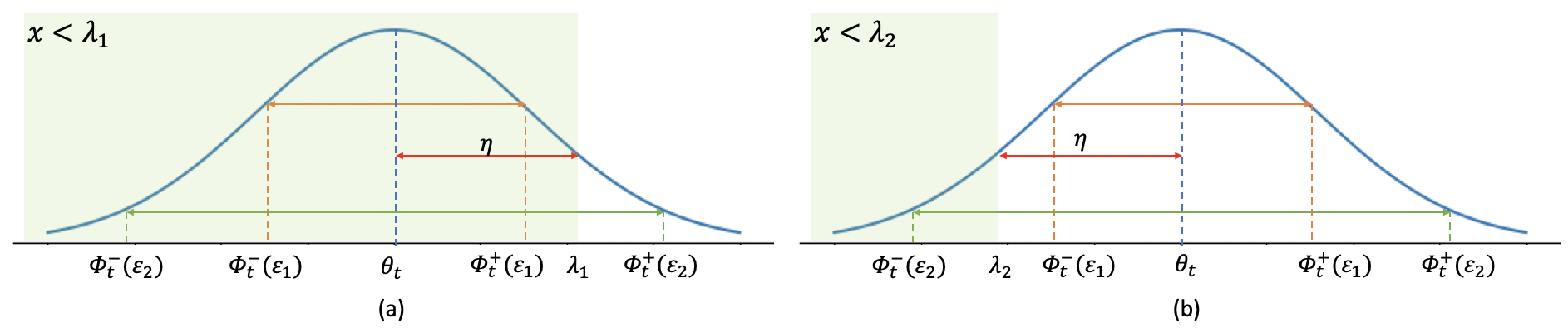}
\caption{Computing confidence guarantees for STL-U formulas $x < \lambda_1$ and $x < \lambda_2$.} 
% \vspace{-0.3cm}
\label{fig:eg_conf}
\end{figure}

In the following, we explain the intuition behind our methods. 
\figref{fig:eg_conf}(a) plots the normal density curve of a Gaussian distribution $\Phi_t$ with the mean $\theta_t$. 
A confidence level $\conflevel$ represents the probability that the corresponding confidence interval $[\lb, \ub]$ contains a target value, calculated as the percentage of the area of the normal density curve. 
When $\conflevel$ approaches $0$, the confidence interval shrinks to a single point $\theta_t$; and when $\conflevel$ approaches $1$, the confidence interval expands to $(-\infty, \infty)$.
In general, the larger the value of $\conflevel$, the wider the confidence interval range. 
For example, \figref{fig:eg_conf}(a) shows confidence intervals for two confidence levels $\conflevel_1$, $\conflevel_2$ and $\conflevel_1 < \conflevel_2$.

To compute the range of confidence levels that guarantee the strong satisfaction of a STL-U formula $\varphi$, 
we first determine the smallest distance $\eta$ between the mean $\theta_t$ and the set of $x$ values that violate $\varphi$.
Any $x$ value within the interval $(\theta_t - \eta, \theta_t + \eta)$ should satisfy $\varphi$.
Thus, we compute the integral $\int_{\theta_t - \eta}^{\theta_t + \eta} \Phi_t(x) dx$ as the upper bound of confidence level guarantee for strong satisfaction, denoted by $\epsilon_s^+$.
Under any confidence level $\conflevel \in (0, \epsilon_s^+)$, the flowpipe is guaranteed to strongly satisfy the STL-U formula $\varphi$.
In the special case when the mean value $\theta_t$ violates $\varphi$, we have $\eta=0$ and $\epsilon_s^+=0$; thus, there does not exist a feasible value of $\conflevel$, under which the flowpipe strongly satisfies $\varphi$.
Consider a STL-U formula $\varphi_1: x < \lambda_1$.
As shown in \figref{fig:eg_conf}(a), the flowpipe under $\conflevel_1$ strongly satisfies $\varphi_1$ because $\conflevel_1 \in (0, \epsilon_s^+)$, 
while the flowpipe under $\conflevel_2$ does not strongly satisfy $\varphi_1$ because $\conflevel_2 > \epsilon_s^+$.

To compute the range of confidence levels that guarantee the weak satisfaction of a STL-U formula $\varphi$, 
we find the smallest distance $\eta$ between the mean $\theta_t$ and the set of $x$ values that satisfy $\varphi$.
We compute the integral $\int_{\theta_t - \eta}^{\theta_t + \eta} \Phi_t(x) dx$ as the lower bound of confidence level guarantee for weak satisfaction, denoted by $\epsilon_w^-$.
Under any confidence level $\conflevel \in (\epsilon_w^-, 1)$, the flowpipe is guaranteed to weakly satisfy the STL-U formula $\varphi$.
When there does not exist any $x$ value satisfying $\varphi$, we have $\eta=\infty$ and $\epsilon_w^-=1$; thus, there does not exist a feasible value of $\conflevel$, under which the flowpipe weakly satisfies $\varphi$.
\figref{fig:eg_conf}(b) shows the same Gaussian distribution $\Phi_t$ as in \figref{fig:eg_conf}(a).
For the STL-U formula $\varphi_2: x < \lambda_2$, the flowpipe under $\conflevel_2$ weakly satisfy $\varphi_2$ because $\conflevel_2 \in (\epsilon_w^-,1)$;
however, the flowpipe under $\conflevel_1$ does not weakly satisfy $\varphi_2$, because $\conflevel_1 < \epsilon_w^-$.

\begin{figure}[t]
\centering
\includegraphics[width=0.95\textwidth]{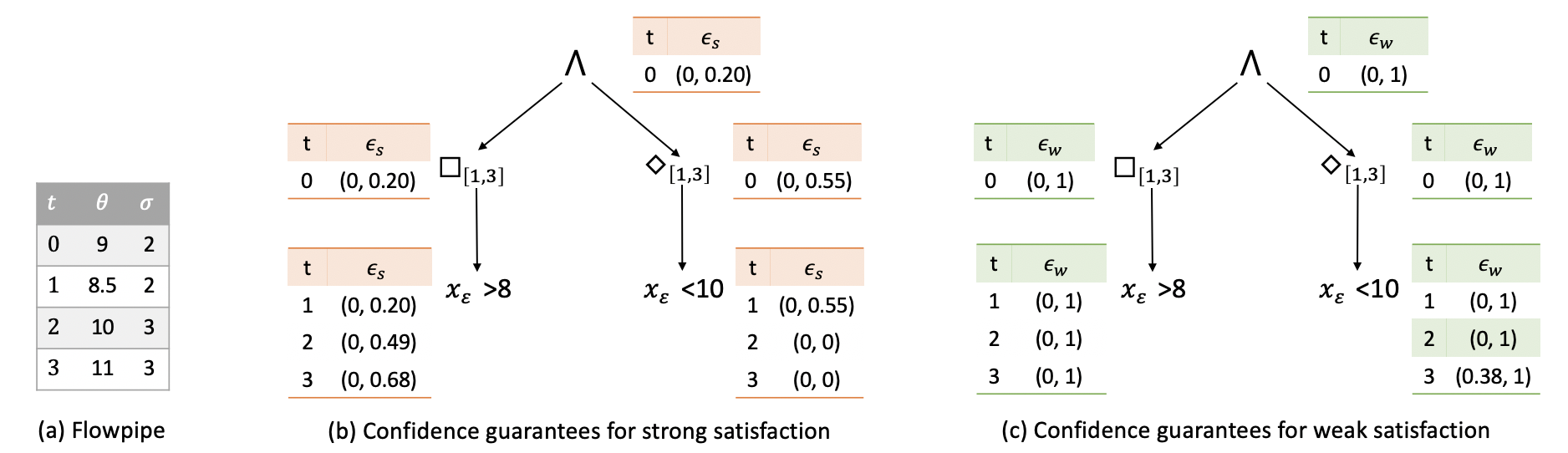}
\caption{Computing confidence guarantees for STL-U formula $\always_{[1,3]}(x_\conflevel>8) \land \eventually_{[1,3]}(x_\conflevel<10)$.} 
% \vspace{-0.3cm}
\label{fig:cf_cfexample}
\end{figure}

We include pseudo code of algorithms for computing confidence guarantees for STL-U strong and weak satisfaction as Algorithm 3 and Algorithm 4 in the Appendix.
Here we use an example to describe the procedure of recursively computing confidence guarantees via parsing the syntax tree of a STL-U formula.
\figref{fig:cf_cfexample}(a) shows an example flowpipe with values of mean $\theta$ and variance $\sigma$ in each time step $t$.
Consider a STL-U formula $\always_{[1,3]}(x_\conflevel>8) \land \eventually_{[1,3]}(x_\conflevel<10)$.
\figref{fig:cf_cfexample}(b) illustrates how to iterate through the formula's syntax tree and compute confidence guarantees for strong satisfaction. 
First, at the left bottom of the tree, we compute the range of confidence levels that can guarantee the strong satisfaction of $x_\conflevel>8$, and obtain the results of $(0, 0.20), (0, 0.49), (0, 0.68)$ for $t \in [1,3]$. 
Then, we move up the tree to compute the confidence guarantee for $\always_{[1,3]}(x_\conflevel>8)$ by taking the intersection of these three ranges, which yields $(0, 0.20)$.
Meanwhile, from the right branch of the tree, we obtain the range of confidence levels that guarantee the strong satisfaction of $x_\conflevel<10$, and taking a union of these ranges for $\eventually_{[1,3]}(x_\conflevel<10)$, which yields $(0, 0.55)$.
At the top of the syntax tree, we take the intersection of $(0, 0.20)$ and $(0, 0.55)$ for $\land$ operation, which yields $(0, 0.20)$ as the final result of confidence guarantees for strongly satisfy the STL-U formula.  
\figref{fig:cf_cfexample}(c) shows a similar process of recursively computing confidence guarantees for weak satisfaction of the same STL-U formula.

\section{Prediction with Logic-Calibrated Uncertainty}
\label{sec:prediction}

Recall from \sectref{sec:motivation} that deterministic prediction models are not suitable to capture the uncertainty exhibited in CPS. 
To address this limitation, we adopt Bayesian RNN models in the proposed predictive monitoring approach.
We describe how to build Bayesian RNN models for prediction and motivate the need for uncertainty calibration in \sectref{sec:schema}.
Then, we define STL-U based criteria for uncertainty calibration in \sectref{sec:criteria}.

%=================================================================================
\subsection{Uncertainty Estimation with Bayesian RNN Models}\label{sec:schema}

\begin{figure}[t]
    \centering
    \includegraphics[width=\textwidth]{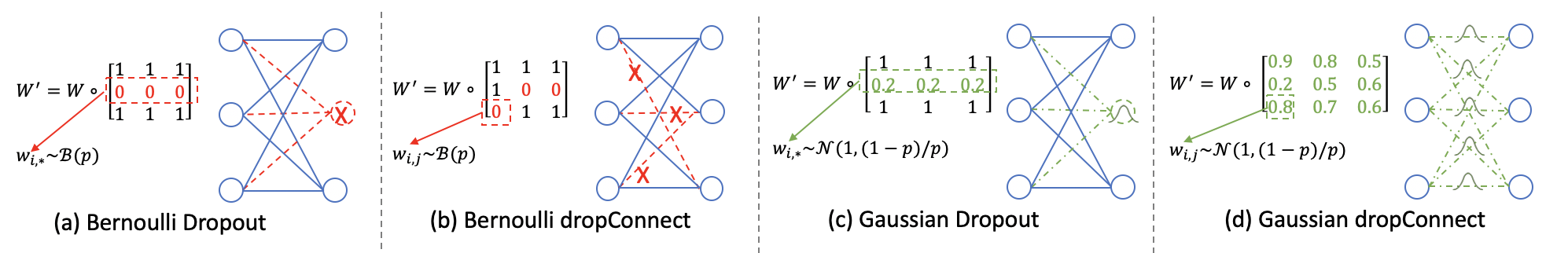}
    \caption{Four commonly used SRTs.}
    \label{fig:dropout}
    \vspace{-0.2cm}
\end{figure}

\begin{figure}
    \centering
    \includegraphics[width=0.75\textwidth]{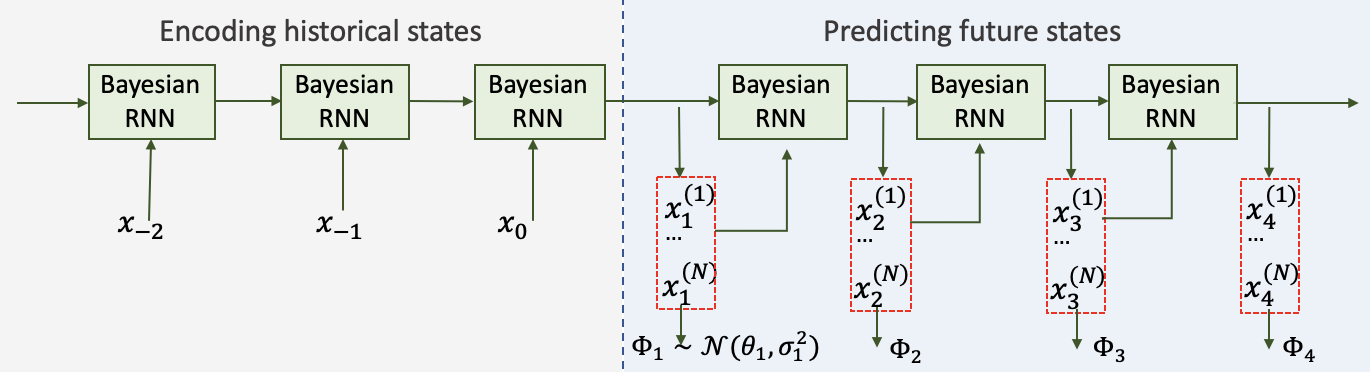}
    \captionof{figure}{Bayesian RNN-based sequential prediction with uncertainty estimation.}
    \label{fig:prediction}
    % \vspace{-0.5cm}
 \end{figure}

Stochastic regularization techniques (SRTs) have been popularly used to cast deterministic deep learning models as Bayesian models for uncertainty estimation~\cite{gal2016uncertainty}. 
Given a well-trained deterministic RNN model with learnable parameters $W$, 
we can obtain a Bayesian RNN model with parameters $W'$ via SRTs that transform $W$ to $W'$
by applying a $n \times n$ mask, where $n$ is the number of neurons in each layer.
Elements of the mask $w$ are sampled from some probability distribution.
The connection from neuron $j$ to neuron $i$ would be dropped if $w_{ij}=0$,
the connection remains the same if $w_{ij}=1$,
and a weight $\beta \in(0,1)$ would be applied to the connection if $w_{ij}=\beta$.
In this work, we consider four commonly used SRTs as illustrated in \figref{fig:dropout}.
Let $p$ denote the dropout rate.  
\begin{itemize}
    \item \emph{Bernoulli dropout}: Each row of the mask is sampled from a Bernoulli distribution, denoted by $w_{i,*} \sim \mathcal{B}(p)$. 
    \item \emph{Bernoulli dropConnect}: Each element of the mask is sampled independently as $w_{i,j} \sim \mathcal{B}(p)$.
    \item \emph{Gaussian dropout}: Each row of the mask is sampled from a Gaussian distribution, denoted by $w_{i,*} \sim \mathcal{N}(1, {(1-p)/p})$.
    \item \emph{Gaussian dropConnect}: Each element of the mask is sampled independently, denoted by $w_{i,j} \sim$ $ \mathcal{N}(1, {(1-p)/p})$. 
\end{itemize}
\figref{fig:prediction} shows how to use the obtained Bayesian RNN model to predict future states based on historical states. 
\revision{We apply the Monte Carlo method to repeat the Bayesian RNN prediction for $N$ times, which yield a set of sequential predictions. 
Thus, we can estimate a Gaussian distribution $\Phi_t\sim \mathcal{N}(\theta_t,\sigma^2_t)$ for each time step $t$, where the mean $\theta_t$ and variance $\sigma_t$ are computed based on the Monte Carlo samples
$\{x^{(1)}_t, \cdots, x^{(N)}_t\}$.} 

\begin{figure}[t]
\centering
\includegraphics[width=\textwidth]{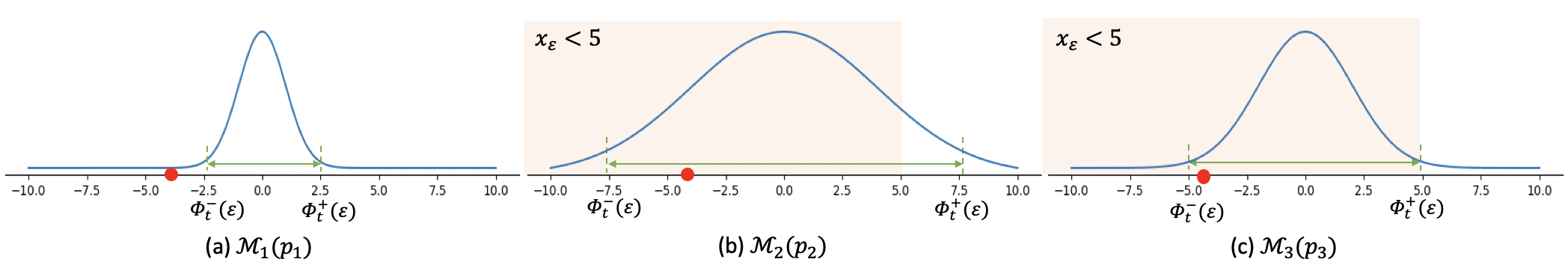}
\caption{Comparison of different uncertainty estimation schemas.}
\vspace{-0.4cm}
\label{fig:creterion}
\end{figure}

Different uncertainty estimation schemas (i.e., SRTs and dropout rates) can yield different uncertainty estimates for the same model trained with the same data. 
How to select the best uncertainty estimation schema for an application still remains an open question. 
Currently, a common practice is to pick a schema empirically, without systematically evaluating how different choices would impact the quality of uncertainty estimates. 
Furthermore, deep learning methods that seek to optimize the prediction accuracy may overestimate the uncertainty. 
For example, consider two distributions predicted with uncertainty estimation schemas $\mathcal{M}_1(p_1)$ and $\mathcal{M}_2(p_2)$, as shown in \figref{fig:creterion}(a) and (b).
$\mathcal{M}_2(p_2)$ is a better schema based on the metric of prediction accuracy, because the target value (red dot) falls within the confidence interval $[\lb, \ub]$ in \figref{fig:creterion}(b) but not in \figref{fig:creterion}(a).
However, $\mathcal{M}_2(p_2)$ yields a higher level of uncertainty, as indicated by the larger confidence interval range. 

In this work, we develop novel criteria that leverage STL-U monitoring results to select uncertainty estimation schemas. 
Here is an example to explain the intuition behind our approach.
Consider two distributions predicted with uncertainty estimation schemas $\mathcal{M}_2(p_2)$ and $\mathcal{M}_3(p_3)$, as shown in \figref{fig:creterion}(b) and (c).
Both distributions fulfill the accuracy metric, because their confidence intervals contain the target value (red dot). 
Suppose that the requirement is to check if a flowpipe strongly satisfies a STL-U formula $x_\varepsilon < 5$.
As shown in \figref{fig:creterion}(c), the distribution predicted with schema $\mathcal{M}_3(p_3)$ strongly satisfies $x_\varepsilon < 5$, because all values in the confidence interval $[\lb, \ub]$ are smaller than 5. 
By contrast, the resulting distribution of schema $\mathcal{M}_2(p_2)$ does not strongly satisfy $x_\varepsilon < 5$, because some values in the confidence interval are greater than 5. 
Thus, based on STL-U monitoring results, we would select $\mathcal{M}_3(p_3)$ rather than $\mathcal{M}_2(p_2)$ as the uncertainty estimation schema, which also yields a tighter bound of estimated uncertainty. 
In the following, we formally define STL-U based criteria for selecting uncertainty estimation schemas.

%In addition, calibrating the uncertainty estimation with system requirements make it clearer to see that if the predictive flowpipe satisfies the requirement, which leads to a higher accuracy on the verification results. It is easier for decision makers to tell if the predictive results satisfy the requirement or not. 

%=================================================================================
\subsection{STL-U Criteria for Uncertainty Calibration}\label{sec:criteria}

\begin{figure}
    \centering
    \includegraphics[width=0.65\textwidth]{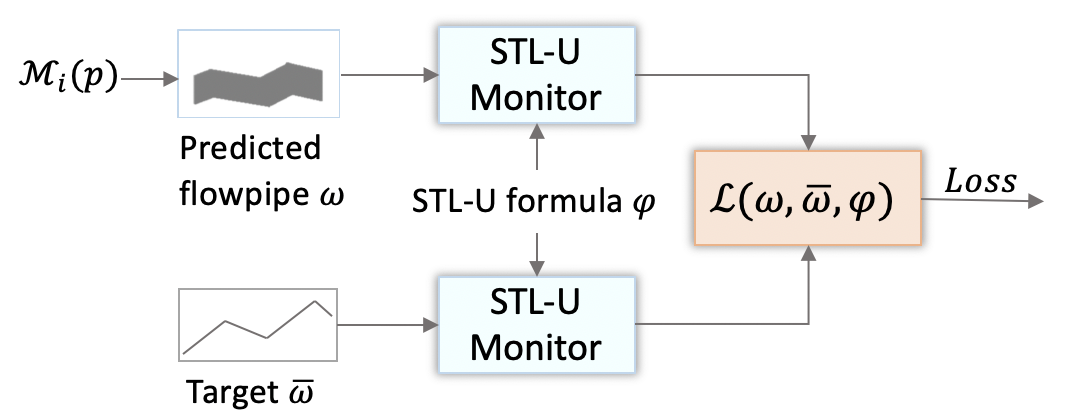}
    \captionof{figure}{STL-U criteria for uncertainty calibration computed as loss functions.}
    \label{fig:criteria}
    \vspace{-0.4cm}
\end{figure}

As shown in \figref{fig:criteria}, given a predicted flowpipe $\omega$ and a target trace $\bar{\omega}$, we can calculate the loss based on monitoring results of $\omega$ and $\bar{\omega}$ with respect to a STL-U formula $\varphi$.
We propose two uncertainty calibration criteria as loss functions based on STL-U satisfaction relations and confidence guarantees, denoted by $\mathcal{L}_\mathsf{sat}$ and $\mathcal{L}_\mathsf{cf}$, respectively. 

\startpara{Criterion based on STL-U satisfaction}
We define $\mathcal{L}_\mathsf{sat}$ based on the linear combination of three functions:
$h_s(\omega, \bar{\omega}, \varphi)$ and $h_w(\omega, \bar{\omega}, \varphi)$
for evaluating if the predicted flowpipe $\omega$ and the target trace $\bar{\omega}$ are consistent in terms of strong and weak satisfaction (or violation) of the STL-U formula $\varphi$, and 
$h_b(\omega, \bar{\omega})$ for evaluating the prediction accuracy by checking if the target trace $\bar{\omega}$ belongs to the predicted flowpipe $\omega$.  
Formally, we define 
 
\begin{equation*}
% \begin{array}{llll}
\begin{split}         
h_{s}(\omega, \bar{\omega}, \varphi)& = \mathbbm{1}\big((\omega \satstrong \varphi \land \bar{\omega} \sat \varphi ) \lor (\omega \notsatstrong \varphi \land \bar{\omega} \notsat \varphi )\big) \\
h_{w}(\omega, \bar{\omega}, \varphi)& = \mathbbm{1}\big((\omega \satweak \varphi \land \bar{\omega} \sat \varphi ) \lor (\omega \notsatweak \varphi \land \bar{\omega} \notsat \varphi )\big) \\
h_b(\omega, \bar{\omega})& =  \mathbbm{1}({ \bar{\omega} \in \omega})\\
\end{split}
% \end{array}
\end{equation*}

\noindent
where $\mathbbm{1}(\phi)$ is an indicator function such that $\mathbbm{1}(\phi) = 1$ if $\phi=\mathsf{True}$, and $\mathbbm{1}(\phi) = 0$ otherwise. 
The loss function is then given by 

\begin{equation*}
    \begin{split}
    \mathcal{L}_\mathsf{sat}(\omega, \bar{\omega}, \varphi)  = 
    1-( \beta_1 \cdot h_s(\omega, \bar{\omega}, \varphi) 
    + \beta_2 \cdot h_w(\omega, \bar{\omega}, \varphi)  
    + (1-\beta_1 -\beta_2) \cdot h_b(\omega, \bar{\omega}))
    \end{split}
\label{eq:qlossl1}
\vspace{-0.5cm}
\end{equation*}

\noindent
where $\beta_1, \beta_2 \in (0,1)$ are real-valued coefficients representing the relative importance of strong/weak satisfaction and prediction accuracy in different domains. 
The goal is to minimize the loss $\mathcal{L}_\mathsf{sat}$, for which we need to maximize the linear combination of $h_s(\omega, \bar{\omega}, \varphi)$, $h_w(\omega, \bar{\omega}, \varphi)$, and $h_b(\omega, \bar{\omega})$.
Intuitively, the higher quality of the prediction in terms of the consistency of STL-U monitoring results and the accuracy compared with the target trace, the lower the loss.

\startpara{Criterion based on STL-U confidence guarantees}
Recall from \sectref{sec:monitor} that, in addition to checking strong/weak satisfaction relations, the STL-U monitor can also compute a range of confidence levels under which the predicted flowpipe is guaranteed to strongly/weakly satisfy a STL-U formula. 
Based on STL-U confidence guarantees, we define the following loss function:
\begin{equation*}
      \mathcal{L}_\mathsf{cf}(\omega, \bar{\omega}, \varphi)  = 
      1 - (\beta_1 \cdot g_s(\omega, \bar{\omega}, \varphi) 
      + \beta_2 \cdot g_w(\omega, \bar{\omega}, \varphi) 
      + (1-\beta_1-\beta_2) \cdot g_b(\omega, \bar{\omega}))
\end{equation*}
where $\beta_1, \beta_2 \in (0,1)$ are real-valued coefficients similar to those used for $\mathcal{L}_\mathsf{sat}$, and $g_s(\omega, \bar{\omega}, \varphi)$, 
$g_w(\omega, \bar{\omega}, \varphi)$ and $g_b(\omega, \bar{\omega})$ are functions defined as follows.

{ 
\begin{equation*}
\begin{split}
    g_s(\omega, \bar{\omega}, \varphi) &=  \begin{cases}
   \epsilon_s^+   & \bar{\omega} \sat \varphi \\
    1 - \epsilon_s^+ & {\bar{\omega}} \notsat \varphi 
    \end{cases}\\
    g_w(\omega, \bar{\omega}, \varphi) &=  \begin{cases}
    1 - \epsilon_w^-  & {\bar{\omega}} \sat \varphi \\
    \epsilon_w^- & {\bar{\omega}} \notsat \varphi 
    \end{cases}\\
    g_b(\omega, \bar{\omega}) &= \inf \left\{\varepsilon \ |\ {\bar{\omega}}_x[t] \in [\Phi_t^-(\varepsilon), \Phi_t^+(\varepsilon)] \mbox{ for all } x \in X, t \in \mathbbm{T}\right\}
     \end{split}
\end{equation*}}

\noindent
where $\epsilon_s^+$ is the upper bound of confidence guarantee for strong satisfaction computed based on \defref{def:strongcl},
$\epsilon_w^-$ is the lower bound of confidence guarantee for weak satisfaction computed based on \defref{def:weakcl},
and $g_b(\omega, \bar{\omega})$ computes the smallest confidence level under which the predicted flowpipe is guaranteed to contain the target trace. 
The goal is to minimize the loss $\mathcal{L}_\mathsf{cf}$, for which we need to maximize the linear combination of $g_s(\omega, \bar{\omega}, \varphi)$, $g_w(\omega, \bar{\omega}, \varphi)$, and $g_b(\omega, \bar{\omega})$.
Intuitively, the lower the loss, the higher quality of predictions in terms of confidence guarantees for strong/weak satisfaction and prediction accuracy.

\startpara{Uncertainty calibration using STL-U criteria}
In order to select the best uncertainty estimation schema, we start with a set of candidate schemas $\uncerF_1(p), \uncerF_2(p), ..., \uncerF_n(p)$.
For each schema with SRT $\uncerF_i$, we tune the dropout rate parameter $p$ using loss functions $\mathcal{L}_\mathsf{sat}$ or $\mathcal{L}_\mathsf{cf}$.
Given a dataset with multiple target traces, we average the losses over all traces to obtain the optimal dropout rate $p^*$.
We compare the losses of candidate schemas equipped with their corresponding optimal dropout rates, and select the best schema $\uncerF^*(p^*)$ that yields the lowest loss. 
Such a process of selecting and turning uncertainty estimation schemas based on STL-U criteria is illustrated as part of \figref{fig:overview}.

We evaluate and compare the performance of different STL-U criteria in \sectref{sec:evaluation}.
Generally speaking, users can choose to use $\mathcal{L}_\mathsf{sat}$ or $\mathcal{L}_\mathsf{cf}$ depending on their needs and problem domains. 
For example, we would recommend applications with strict safety requirements (e.g., a fire risk prediction and control service) to adopt $\mathcal{L}_\mathsf{sat}$ for checking strong satisfaction relations. 
By contrast, $\mathcal{L}_\mathsf{cf}$ is more flexible and does not require a pre-defined confidence level, which is suitable for applications that do not have a specific confidence level yet try to optimize the uncertainty estimation (e.g., a newly deployed energy control service).

\section{Evaluation}
\label{sec:evaluation}

We conducted experiments to evaluate the proposed approach.
In \sectref{sec:exp-predication}, we compare STL-U criteria for uncertainty calibration with state-of-the-art baselines using real-world CPS datasets.
In \sectref{sec:exp-monitor}, we demonstrate the performance of our approach on real-time predictive monitoring in a simulated smart city case study.
The experiments were run on a machine with 2.2GHz CPU, \revision{32GB memory}, and Nvidia GeForce RTX 2080Ti GPU. %The operating system is Centos 7.

%=================================================================================  
\subsection{Evaluating STL-U Criteria for Uncertainty Calibration}\label{sec:exp-predication}
We use two real-world city datasets (i.e., air quality and traffic volume datasets) described in \sectref{sec:motivation}. We split each dataset into 80\% data for RNN training, 10\% data for STL-U based uncertainty estimation \revision{(i.e., tuning Bayesian RNN)}, and 10\% data for testing. \revision{We trained the model for 30 epochs.} 
% In the following experiments, we assume the time domain can be represented by $\mathbb{T} = \{1,2, \dots,T\}$.} 

% the deep learning prediction generates 130,000 and 51,350 flowpipes for the air quality case and the traffic case, respectively. 
% We monitor 390 and 4,470 STL-U requirements for the air quality case and the traffic case, respectively. 

\begin{figure}[t]
\centering
\includegraphics[width=\textwidth]{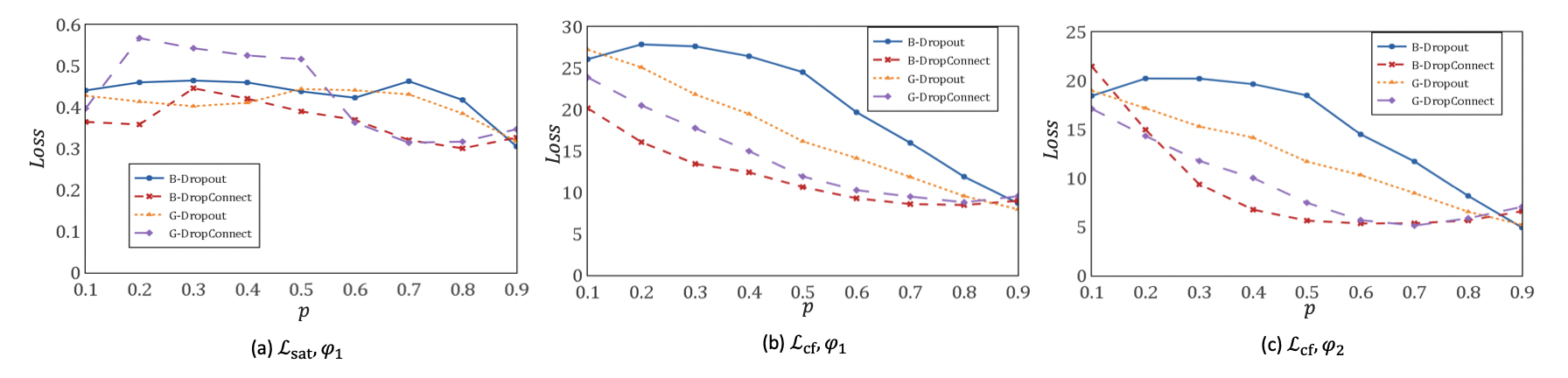}
\caption{Selecting uncertainty estimation schemas using different STL-U criteria.} 
\vspace{-0.2cm}
\label{fig:rq1}
\end{figure}

\startpara{Comparing different STL-U criteria}
\figref{fig:rq1} plots the loss obtained using different STL-U criteria when varying uncertainty estimation schemas (i.e., SRT and dropout rate $p$) for the air quality dataset. We trained an LSTM as the underlying RNN model.
\figref{fig:rq1}(a) shows the results of using STL-U criterion $\mathcal{L}_\mathsf{sat}$ for $\varphi_1=\always_I(x_\varepsilon < \lambda)$,
where the schema of Bernoulli DropConnect with $p=0.8$ yields the lowest loss. 
\figref{fig:rq1}(b) shows the results of using STL-U criterion $\mathcal{L}_\mathsf{cf}$ for $\varphi_1$,
where the schema of Gaussian Dropout with $p=0.9$ yields the lowest loss.
\figref{fig:rq1}(c) shows the results of using STL-U criterion $\mathcal{L}_\mathsf{cf}$ for $\varphi_2=\eventually_I(x_\varepsilon < \lambda)$,
where the schema of Bernoulli Dropout with $p=0.9$ yields the lowest loss.
Thus, the optimal uncertainty estimation schema varies based on different STL-U criteria.
The experiments demonstrate that the proposed approach is feasible for the automated selection of optimal schemas based on system requirements and user demands (i.e., whether the user is interested in checking requirement satisfaction or computing confidence guarantees).

\begin{table*}[t]
\centering
\scriptsize
\caption{Results of comparing STL-U criteria with six baselines.}
\begin{tabular}{l|ccccc|ccccc}
\toprule
& \multicolumn{5}{c}{\textbf{Air Quality}}& \multicolumn{5}{c}{\textbf{Traffic Volume}}\\
     \textbf{Criteria}                &	\textbf{SRT}	&	$p$	&	\textbf{HeterLoss}	&	\textbf{Accuracy}	&	\textbf{F1-Sat}      &	\textbf{SRT}	&	$p$	&	\textbf{HeterLoss}	&	\textbf{Accuracy}	&	\textbf{F1-Sat} \\
                      \cmidrule{1-11}
-	&	B-Dropout	&	0.81	&	183.9	&	0.67	&	0.34	&	B-Dropout	&	0.50	&	0.63	&	0.79	&	0.17	\\
-	&	B-DropConnect	&	0.53	&	121.0	&	0.69	&	0.22	&		B-DropConnect	&	0.74	&	{0.23}	&	0.38	&	0.51	\\
-	&	G-Dropout	&	0.45	&	152.8	&	0.76	&	0.10	&		G-Dropout	&	0.50	&	0.66	&	0.79	&	0.17		\\
-	&	G-DropConnect	&	0.58	&	129.4	&	0.78	&	0.12	&		G-DropConnect	&	0.54	&	0.25	&	0.56	&	0.44		\\
% ConcreteD~\cite{gal2017concrete}	&	-	&	-	&	128.6	&	{0.91}	&	0.69	&	-	&	-	&	2.96	&	{1.00}	&	0.11	\\
\cmidrule{1-1} \cmidrule{2-2} \cmidrule{3-11}
$\mathcal{L}_\mathsf{acc}$	&	B-DropConnect	&	0.53	&	121.0	&	0.69	&	0.22	&		B-DropConnect	&	0.74	&	0.23	&	0.38	&	0.51	\\
$\mathcal{L}_\mathsf{ht}$	&	G-Dropout	&	0.50	&	{119.2}	&	{0.81}	&	0.65	&	G-Dropout	&	0.50	&	0.66	&	0.79	&	0.17		\\
$\mathcal{L}_\mathsf{sat}$	&	G-DropConnect	&	{0.81}	&	154.1	&	0.80	&	\textbf{0.81} 	&	B-DropConnect	&	0.58	&	0.24	&	0.51	&	\textbf{{0.67}}		\\
$\mathcal{L}_\mathsf{cf}$	&	B-DropConnect	&	0.73	&	165.4	&	0.79	&	\textbf{0.76}	&	B-Dropout	&	{0.90}	&	0.3	&	0.78	&	\textbf{0.68}	\\
\bottomrule
\end{tabular}
\label{tab:diffcriteria}
\end{table*}

\startpara{Comparing STL-U criteria with baselines}
\tabref{tab:diffcriteria} shows the results of applying uncertainty estimation with STL-U criteria (bottom two rows) and state-of-the-art baselines (top six rows) to the testing data of air quality and traffic volume datasets. We trained an LSTM as the underlying RNN model for each dataset. 
We consider six \emph{baselines} for comparison.
The top four rows of the table are results of using four SRTs with optimal dropout rates $p$ tuned based on the prediction accuracy (i.e., the percentage of target traces covered in the predicted flowpipes).
The next two rows are results based on optimizing the uncertainty estimation schema using two commonly used criteria: $\mathcal{L}_\mathsf{acc}$ is the loss function concerning the F1-score of prediction accuracy (i.e., if the target trace is covered by the predicted flowpipe), 
and $\mathcal{L}_\mathsf{ht}$ is the loss function approximating the Heteroscedastic aleatoric uncertainty~\cite{kendall2017uncertainties}. \revision{For the hyperparameters in loss functions, we use $\beta_1 = 0.2$, $\beta_2 = 0.2$ for $\mathcal{L}_{sat}$, and $\beta_1 = 0.3$, $\beta_2 = 0.3$ for $\mathcal{L}_{cf}$.}

\revision{We compare their performance in terms of three \emph{metrics} shown in columns of the table:
\begin{itemize}
    \item Heteroscedastic loss: $\mathsf{HeterLoss} = \frac{1}{MT} \sum_{i=1}^M \sum_{t=1}^T (\frac{||y_t^{(i)} - \theta_t^{(i)}||^2}{2{(\sigma_t^{(i)})}^2} + \frac{1}{2}\log 2{\sigma_t^{(i)}})$, where $M$ represents the total number of instances in the testing data and $T$ represents the length of the predicted sequence;
    \item Prediction accuracy (RMSE): $\mathsf{Accuracy} = \frac{1}{MT}  \sum_{i=1}^M \sum_{t=1}^T \frac{||y_t^{(i)} - \theta_t^{(i)}||^2}{2{(\sigma_t^{(i)})}^2}$.
    \item F1-score comparing the STL-U requirement satisfaction for the predicted and target sequences: $\mathsf{F1}$-$\mathsf{Sat} = \frac{TP}{TP + \frac{1}{2} (FP + FN)}$, where $TP, FP, FN$ represents number of true positives, number of false positives, and number of false negatives, respectively. 
\end{itemize}}
% (1) 
% (2) 
% and (3) 
%and (3) F1-score of whether the predicted flowpipes having the same STL-U monitoring results with the target traces for the satisfaction of requirement $(x_\conflevel<\threshold_1)\mathcal{U}_I(x_\conflevel<\threshold_2)$.
The results show that both STL-U criteria yield significant higher F1-scores of requirement satisfaction than all six baselines, which having comparable performance with baselines in terms of Heteroscedastic loss and accuracy.
Low F1-scores of requirement satisfaction indicate that flowpipes predicted using baselines can be barely used for monitoring city requirements due to the low quality of estimated uncertainty (i.e., the predicted flowpipes may contain too much noise to obtain meaningful results about requirement violations).
Thus, using STL-U criteria to calibrate the uncertainty estimation is an essential step for the predictive monitoring.

\begin{figure}[t]
\centering
\includegraphics[width=13.6cm]{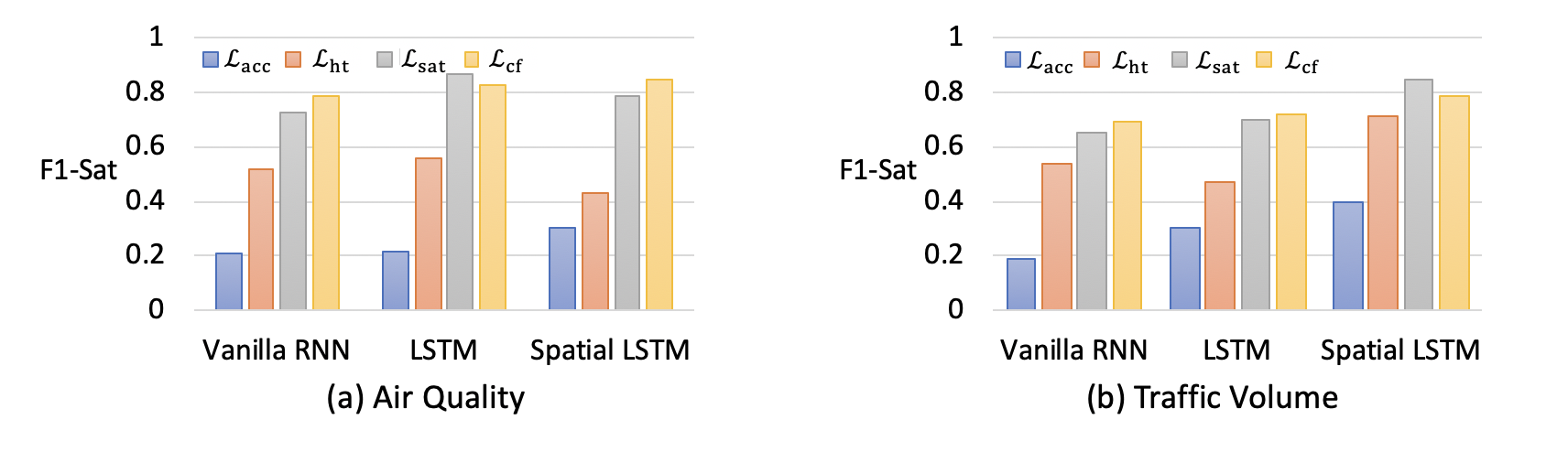}
\vspace{-0.2cm}
\caption{Results of comparing different RNN models.} 
\label{fig:rq3}
\vspace{-0.6cm}
\end{figure}

\startpara{Comparing different RNN models}
\figref{fig:rq3} compares the F1-score of requirement satisfaction of applying different uncertainty calibration criteria on three types of RNN models: (1) Vanilla RNN, (2) LSTM, and (3) Spatial LSTM~\cite{kong2018hst}.
The results show that STL-U criteria $\mathcal{L}_\mathsf{sat}$ and $\mathcal{L}_\mathsf{cf}$ significantly outperform baseline criteria $\mathcal{L}_\mathsf{acc}$ and $\mathcal{L}_\mathsf{ht}$ across all three RNN models for both datasets. 
In addition, both STL-U criteria yield comparable performance across different RNN models. 
Using $\mathcal{L}_\mathsf{sat}$ with an LSTM model and a Spatial LSTM model result in the highest F1-score for the air quality dataset and the traffic volume dataset, respectively. 
The experiments demonstrate that our proposed approach of uncertainty estimation and calibration is compatible with different underlying RNN models.

%=================================================================================  
\subsection{Real-time Predictive Monitoring for a Simulated Smart City}
\label{sec:exp-monitor}

We set up a \revision{closed-loop } simulated smart city based on the \emph{Simulation of Urban MObility} (SUMO) platform~\cite{sumo} using real-world data of New York City~\cite{nycopendata}.
We implemented ten smart services in the simulated smart city, including S1: Traffic Service, S2: Emergency Service, S3: Accident Service, S4: Infrastructure Service, S5: Pedestrian Service, S6: Air Pollution Control Service, S7: PM2.5/PM10 Service, S8: Parking Service, S9: Noise Control Service, and S10: Event Service. 
We built a prototype implementation of the STL-U based predictive monitoring and applied it for the predictive monitoring of 390 requirements concerning different city performance metrics (e.g., AQI, traffic volume, noise) in various locations of the simulated smart city.
When a smart service requests an action, we predict future city states under the influence of the requested action and monitor if city requirements would be violated. 
\revision{Based on the real-time predictive monitoring results generated by our approach, the control center can decide if the requested action should be accepted or rejected to prevent any potential requirement violation. For the details of the decision making process, we follow the methods in CityResolver~\cite{ma2018cityresolver}, which is a decision making system for conflict detection and resolution in smart cities. 
As an intuitive example, when a smart navigation service requests an action to direct vehicles to a school area to release traffic congestion, the predictive monitoring approach first predicts the future sequences of noise levels and air pollution levels, and verifies them with STL-U specified city requirements on noise and air pollution in the school areas. If the predicted sequences satisfy the requirements, the requested action will be approved; however, if they violate the requirements, the control center will generate a resolution similar to CityResolver.   
In our experiment, we run the simulated New York city with STL-U predictive monitoring for 30 simulation days and obtain the results regarding the metrics in \tabref{tab:performance}. } 
Experimental results show that our approach is efficient in handling a large number of flowpipes and requirements. 
\revision{We did not include the execution time of experiments, because the implementation of our prototype tool is not optimized yet. However,
it only takes about 281 seconds to check the satisfaction of 130,000 flowpipes that predict AQI in eight future time units. }

\begin{table}[t]
\caption{Results of comparing the impact of STL-U based predictive monitoring with two baselines.}
\label{tab:performance}
\tablefontsize
% \small
\begin{tabular}{l|c c c}
\toprule
\textbf{ City Performance Metrics}  & \textbf{No Monitor}   & \textbf{LSTM + STL Monitor} & \textbf{STL-U Predictive Monitor} \\ \hline
{Number of Violation}         & - & 267         &  \textbf{189}           \\ 
{Air Quality Index}                     & 68   & 57       & \textbf{43 }         \\ 
{Noise (db) }                 & 73   & 49       & \textbf{48 }         \\ 
{Emergency Waiting Time (s)} & 20   & 14       & \textbf{10 }        \\ 
{Vehicle Waiting Number }    & 22    & 18        & \textbf{15}          \\ 
{Pedestrian Waiting Time (s)}& 190   & 148       & \textbf{121}          \\ 
{Vehicle Waiting Time (s)}  & 112  & 90       & \textbf{80}         \\ 
\bottomrule
\end{tabular}
\vspace{-0.6em}
\end{table}

\tabref{tab:performance} compares STL-U based predictive monitoring's impact on city performance with two \emph{baselines}: (1) running the simulated city without predictive monitoring, and (2) running the simulated city with a basic predictive monitoring component implemented with a deterministic LSTM predictor and a STL monitor. 
The results are based on 30-day data in the simulated city.
First, we observe that our predictive monitor detects fewer requirement violations for the predicted future city states than the baseline method (2).
This is because our approach uses a Bayesian LSTM predictor with calibrated uncertainty, which can generate more accurate predictions about future city states than the deterministic predictor, and thus reducing the number of spurious violations. 
Furthermore, the results show that our approach has the potential to improve various city performance metrics.
For example, compared with the two baselines, our approach reduces the air quality index by 36.8\% and 24.6\%, and reduces the emergency vehicle waiting time by 50\% and 28.6\% in the simulated city.

\section{Related Work}
\label{sec:related}
% 30 
 
\startpara{Predictive monitoring for CPS}
The research area of predictive monitoring has been drawing increasing attention in recent years. 
For example, (Bayesian) Neural Predictive Monitoring~\cite{bortolussi2019neural,BortolussiCPSS20} checks predictions about neural state classification and uses a principled criteria to reject predictions that are likely to be incorrect;
a predictive monitor for rare failures is developed in~\cite{babaee2019accelerated} using Discrete-Time Markov Chains trained with samples of rare events; 
\revision{STLnet~\cite{ma2020stlnet} incorporates predictive monitoring into the learning process and enhance RNN-based sequential prediction models to follow STL specified model properties in both training and testing processes. }
Prevent~\cite{babaee2018predictive} and other runtime verification techniques with state estimation~\cite{StollerBSGHSZ11,KalajdzicBSSG13,BartocciGKSSZS12} use Hidden Markov Models or Dynamic Bayesian Networks to learn and predict the probability of a hidden state satisfying a safety property. \revision{A more recent approach~\cite{WagaAH21} uses instead linear hybrid models of the system under monitoring to bound the uncertainty in the gaps between consecutive samples.}

These existing works mostly focus on monitoring individual predictions rather than sequential predictions.
A more recent work~\cite{qin2020clairvoyant} applies statistical time-series analysis techniques (e.g., ARIMA) to forecast future signal values and computes the satisfaction probability of a STL formula over the prediction horizon; however, the applicability of this approach is limited by the assumption that a joint probability distribution of predictions over multiple time-points can be estimated.
By contrast, our approach considers uncertain sequential predictions generated by Bayesian RNN models, which are generally applicable to many CPS domains.

\startpara{Temporal logic based runtime monitoring}
Over the past decades, tremendous progress has been made in developing techniques and tools of runtime monitoring (also called runtime verification) based on rigorous specifications expressed in various temporal logics (e.g., LTL, STL).
For example, a survey on STL-based runtime monitoring for CPS is provided in~\cite{chapter5}, which includes applications such as automotive systems and medical devices;
a STL-based framework is developed in~\cite{ma2018cityresolver} for detecting requirement violations in smart cities; 
SaSTL~\cite{ma2020sastl} extends STL for runtime monitoring of spatial-temporal properties in CPS;
and another spatial-temporal logic named SpaTeL is applied to monitor the power grid in~\cite{spatel}. 
However, most of the literature focuses on monitoring deterministic multi-variable signals,
which is a limiting factor when we need to monitor predictive models and to reason about uncertainty.

There are some attempts to handle uncertainty by incorporating random variables in predicates.
For example, C2TL~\cite{JhaRSS18} checks the probability of a deterministic signal satisfying a linear constraint whose coefficients are random variables;
PrSTL~\cite{SadighK16} uses atomic predicates that are parameterized with a time-varying random variable over a deterministic signal;
StSTL~\cite{STSTL} checks the probability of a real-valued measurable function over stochastic signals; and StTL~\cite{STSTL2} extends StSTL to reason about the robustness of requirement satisfaction.  
Our approach differs from these previous works in several aspects.
First, the proposed STL-U monitor checks a flowpipe signal that contains an infinite set of uncertain sequences rather than a single sequence. 
Moreover, instead of computing a single probability value of satisfying a predicate, 
STL-U reasons about the uncertainty captured by confidence intervals of the flowpipe signal, which is a more suitable representation of uncertainty estimated from Bayesian deep learning.

The problem of monitoring an infinite set of sequences has been studied before in~\cite{roehm2016stl}
for the reachability analysis of continuous and hybrid system models.
This work proposes a Reachset Temporal Logic (RTL), which extends STL and is defined on the reach sequence (i.e., a function mapping time to the set of states reachable from a set of initial states and uncertain inputs).
The notion of reachability in RTL (i.e., checking if all the values within the reach sequence satisfies a formula) can be encoded as STL-U strong satisfaction with our approach.
\revision{In RTL, the formulas are in positive normal form: the negation operator can appear only in basic propositions while it remains undefined for generic formulas that may include also temporal operators.  Furthermore, the RTL-based model checking algorithm is limited to a specific fragment of RTL where 
the only possible temporal operator that can
be used is the \emph{next} operator. 
Similar to RTL~\cite{roehm2016stl}, the parameter synthesis approach presented in~\cite{DangDP15} introduces a new semantics for the positive normal form fragment of STL that is defined on sets of traces rather than on a single trace. 

In our approach, we introduce the notion of strong/weak semantics to handle the negation operator with more generic formulas: we define the evaluation of strong satisfaction for a formula $\neg \varphi$ (for both atomic predicates and temporal formulas) as equivalent 
to the violation of the weak satisfaction for the formula $\varphi$ and vice-versa.  Our work takes inspiration from the paper of Eisner et al.~\cite{EisnerFHLMC03} where the authors propose a weak/strong semantics to reason with linear temporal logic (LTL) on truncated paths: the weak semantics provides an optimistic view of the satisfaction of an LTL formula on a truncated path, while the strong semantics provides a pessimistic view. In our approach instead, the weak/strong semantics is 
used to change the existential/universal quantifier 
when we interpret a proposition 
over a confidence interval.}

%EisnerFHLMC03

\startpara{Uncertainty estimation in deep learning}
While most deep learning models do not offer the uncertainty of their predictions~\cite{gal2016uncertainty}, works that capture the uncertainty (or confidence) of the prediction can be dated to the early development of neural networks in the 90s'. Bayesian Neural Network~\cite{mackay1992practical} represents a probabilistic model that infers a distribution as output. It is known to be robust and resilient to overfitting. However, the hardness of inference prevents the prevalence of the model in practice. Following these directions, several works ~\cite{hinton1993keeping, graves2011practical} use variational inference to perform an approximated inference on Bayesian Neural Networks. 
% The recent development of Bayesian Neural Networks include~\cite{louizos2016structured, sun2019functional, zhang2018noisy, shi2018spectral}.
Aside from variational inference,  Monte Carlo Dropout is another approach to obtain uncertainty estimation of the model~\cite{gal2016dropout,  zhu2017deep}. By exploiting the dropout structure in the deep neural network, these approaches turn the original Neural Network model into a simple Bayesian Neural Network without changing the structure and apply approximated inference with the Monte Carlo approach. 
Existing works~\cite{xiao2019quantifying, zhu2017deep, kendall2017uncertainties} mostly focus uncertain estimation on single-time classification or regression tasks. This paper focuses on the case of time series prediction. Moreover, in contrast to previous measures of uncertainty that are rather empirical, our work proposes a formal framework to model and define requirements to the output distribution. Our work can thus be used to provide a confidence guarantee of the model prediction and to evaluate the quality of the uncertainty estimation.

\section{Conclusion}
\label{sec:conclusion}

We developed a novel predictive monitoring approach for CPS, which consists of a logic-calibrated Bayesian RNN prediction model that continuously generates sequential predictions of future states, 
and a novel STL-U monitor that checks if the generated predictions satisfy CPS requirements.
Additionally, we proposed novel criteria based on STL-U monitoring results to calibrate uncertainty estimation in Bayesian deep learning for the predictive monitor.  
The experimental results show that STL-U criteria leads to improved uncertainty estimation in various Bayesian deep learning models, and STL-U based predictive monitor significantly improves performance metrics in a simulated smart city study. 

The proposed STL-U monitor is generally applicable for monitoring an infinite set of sequences beyond those generated by Bayesian deep learning. For example, STL-U monitor can also check trajectories of continuous and hybrid systems (e.g., those considered in \cite{roehm2016stl}). 
In addition, the proposed STL-U criteria for uncertainty calibration can be used in a broad spectrum of deep learning applications. 
As demonstrated in \sectref{sec:evaluation}, STL-U criteria can be used for the automated selection of optimal uncertainty estimation schemas and are compatible with different types of RNN models. 
Applying STL-U criteria for uncertainty calibration does not require knowledge about the inner working of deep learning models and stochastic regularization techniques.
Thus, they are amendable for different deep learning applications.  

% \linelabel{l:1}
There are several directions to explore for the future work. 
First, we will extend STL-U logic with quantitative semantics (e.g., robustness of requirement satisfaction).
Second, we will explore the theoretical implications of STL-U criteria on uncertainty calibration for Bayesian deep learning. 
\revision{Third, we will investigate the scalability and efficiency of proposed STL-U monitoring algorithms for more complex specifications (e.g., those with multiple layers of nesting temporal operators). 
Last but not least, we will apply it to a wide range of real-world CPS applications such as autonomous driving, smart health, and smart homes. }

%First, our approach is orthogonal of the structure of the underlying sequential prediction models. In this paper, we presented performance improvements when applying to vanilla RNN, LSTM and Spatial LSTM. It can also adapt to other sequential prediction models easily. Secondly, STL-U criteria select and tune the SRTs based on the system requirements and results of predictive monitoring, it does not require any knowledge or revision on the SRTs. The advantage of treating the deep learning models and SRTs as a black box is that it has the potential to support state-of-the-art or newly developed models or SRTs for uncertainty calibration in the future. 
%Compared with traditional uncertainty estimation methods, the proposed logic-based solution can lead to more flexible and application-customized uncertainty calibration for sequential prediction tasks.

% In the paper, STL-U criteria applies system requirements to calibrate uncertainty. It increases the F1 score of predictive monitoring of these requirements significantly comparing to the traditional uncertainty estimation approaches. From our observation, we found that it also improves the F1 score of of predictive monitoring of other requirements which are not used in the training process. It indicates that our approach improves the quality of uncertainty estimation in general and not only subject to the training requirements. However, we did not prove it in theory. In the future work, we will systematically investigate the influence on a larger scale of untrained requirements experimentally and theoretically. 

\bibliographystyle{ACM-Reference-Format}
\bibliography{ref_final.bib}

% \clearpage
\vspace{-0.2cm}
\section*{Appendix}

\revision{
% \begin{theorem}[Soundness]\label{th:correct}
%  If $(\traceset, t)$ strongly satisfies a STL-U formula $\varphi$, then it will strongly satisfy $\varphi$; If $(\traceset, t)$ does not weakly satisfy a STL-U formula $\varphi$, then it will not strongly satisfy $\varphi$. Formally, we have, 
% \begin{equation*}
% \begin{array}{ll}
% (\traceset, t) \satstrong \varphi &\Rightarrow  (\traceset, t) \satweak \varphi \\
% (\traceset, t) \notsatweak \varphi & \Rightarrow  (\traceset, t) \notsatstrong \varphi \\
% \end{array}
% \end{equation*}

% \end{theorem}

\begin{proof}[Proof of \thref{th:correct}]
We just need to prove $(\traceset, t) \satstrong \varphi \Rightarrow  (\traceset, t) \satweak \varphi$ by structural induction below. By contraposition, we have $((\traceset, t) \notsatweak \varphi  \Rightarrow  (\traceset, t) \notsatstrong \varphi) \eqdef ((\traceset, t) \satstrong \varphi \Rightarrow  (\traceset, t) \satweak \varphi)$.
\begin{itemize}
    \item Base case $\mu_\mathsf{x}$:  
    by \defref{def:strongboolean}, $(\traceset, t) \satstrong \varphi \eqdef \forall x \in [\lb,\ub], f(x)>0$, it indicates that $\exists x \in [\lb,\ub], f(x)>0 \eqdef (\traceset, t) \satweak \varphi$.
    \item Inductive case $\neg \varphi$: from inductive hypothesis $(\traceset, t) \satstrong \varphi \Rightarrow (\traceset, t) \satweak \varphi$ (and consequently $((\traceset, t) \notsatweak \varphi  \Rightarrow  (\traceset, t) \notsatstrong \varphi)$, we need to prove that $(\traceset, t) \satstrong \neg \varphi \Rightarrow (\traceset, t) \satweak \neg \varphi$.
    
    We have $(\traceset, t) \satstrong \neg \varphi \eqdef (\traceset, t) \notsatweak \varphi \Rightarrow (\traceset, t) \notsatstrong \varphi \eqdef
    (\traceset, t) \satweak \neg \varphi$.
    
    \item Inductive case $\varphi_1 \land \varphi_2$: from inductive hypothesis $(\traceset, t) \satstrong \varphi_1 \Rightarrow (\traceset, t) \satweak \varphi_1$ and $(\traceset, t) \satstrong \varphi_2 \Rightarrow (\traceset, t) \satweak \varphi_2$, we need to prove that $(\traceset, t) \satstrong \varphi_1 \land \varphi_2 \Rightarrow (\traceset, t) \satweak \varphi_1 \land \varphi_2$.
    
    By \defref{def:strongboolean} and \defref{def:weakboolean}, we have $(\traceset, t) \satstrong \varphi_1 \land \varphi_2 \eqdef
    (\traceset, t) \satstrong \varphi_1 \land (\traceset, t) \satstrong \varphi_2 \Rightarrow (\traceset, t) \satweak \varphi_1 \land (\traceset, t) \satweak \varphi_2
    \eqdef (\traceset, t) \satweak \varphi_1 \land \varphi_2$.
    
    \item Inductive case $\always_I\varphi$: from inductive hypothesis $(\traceset, t) \satstrong \varphi \Rightarrow (\traceset, t) \satweak \varphi$, we need to prove that $(\traceset, t) \satstrong \always_I \varphi \Rightarrow (\traceset, t) \satweak  \always_I \varphi$. 
    
    By \defref{def:strongboolean} and \defref{def:weakboolean}, we have $(\traceset, t) \satstrong \always_I \varphi \eqdef \forall t' \in (t+I), (\traceset, t') \satstrong \varphi$, which indicates that $\forall t' \in (t+I), (\traceset, t') \satweak \varphi \eqdef (\traceset, t) \satweak  \always_I \varphi$.
    
    \item Inductive case $\eventually_I\varphi$: from inductive hypothesis $(\traceset, t) \satstrong \varphi \Rightarrow (\traceset, t) \satweak \varphi$, we need to prove that $(\traceset, t) \satstrong \eventually_I \varphi \Rightarrow (\traceset, t) \satweak  \eventually_I  \varphi$.
    
    By \defref{def:strongboolean} and \defref{def:weakboolean}, we have $(\traceset, t) \satstrong \eventually_I \varphi \eqdef \exists t' \in (t+I), (\traceset, t') \satstrong \varphi$, which indicates that $\exists t' \in (t+I), (\traceset, t') \satweak \varphi \eqdef (\traceset, t) \satweak  \eventually_I \varphi$.
    
    \item Inductive case $\varphi_1\until\varphi_2$: from inductive hypothesis $(\traceset, t) \satstrong \varphi_1 \Rightarrow (\traceset, t) \satweak \varphi_1$ and $(\traceset, t) \satstrong \varphi_2 \Rightarrow (\traceset, t) \satweak \varphi_2$, we prove that $(\traceset, t) \satstrong \varphi_1 \until \varphi_2 \Rightarrow (\traceset, t) \satweak \varphi_1 \until \varphi_2$.
    
    By \defref{def:strongboolean} and \defref{def:weakboolean}, we have $(\traceset, t) \satstrong \varphi_1 \until \varphi_2 \eqdef \exists t' \in (t+I) \cap \Tset, (\traceset, t') \satstrong \varphi_2    \mbox{ and }  
         \forall t'' \in (t, t'), (\traceset, t'') \satstrong \varphi_1$, which indicates that $\exists t' \in (t+I) \cap \Tset, (\traceset, t') \satweak \varphi_2    \mbox{ and }  
         \forall t'' \in (t, t'), (\traceset, t'') \satweak \varphi_1$, it is equivalent to $(\traceset, t) \satweak \varphi_1 \until \varphi_2$.
    
\end{itemize}
 
\end{proof}
}
\revision{
\paragraph{Additional properties and examples}
\label{pro:neg}
By applying the rules of the weak/strong semantics for negation we have that the following properties hold:
\begin{equation*}
\begin{array}{lclclcl}
(\traceset, t) \satstrong \neg \neg \varphi &\eqdef &  (\traceset, t) \satstrong \varphi & &
(\traceset, t) \satweak \neg \neg \varphi &\eqdef & (\traceset, t) \satweak \varphi \\
\end{array}
\end{equation*}

\noindent By applying the De Morgan's laws we have the following: 
\begin{equation*}
\begin{array}{lclcl}
(\traceset, t) \notsatweak \neg \varphi_1 \land (\traceset, t) \notsatweak \neg \varphi_2 & \eqdef & \neg ( (\traceset, t) \satweak (\neg \varphi_1 \lor \neg \varphi_2)) & \eqdef & 
(\traceset, t) \notsatweak \neg (\varphi_1 \land \varphi_2) \\
(\traceset, t) \notsatstrong \neg \varphi_1 \land (\traceset, t) \notsatstrong \neg \varphi_2 & \eqdef &  \neg ( (\traceset, t) \satstrong (\neg \varphi_1 \lor \neg \varphi_2)) & \eqdef & 
(\traceset, t) \notsatstrong \neg (\varphi_1 \land \varphi_2) \\
\end{array}
\end{equation*}
\noindent Furthermore, to clarify the duality of the two semantics, we can interpret the weak semantics using the interval intersection and the strong semantics using the interval inclusion. For example, let us define as basic propositions: $\mu_x^{f}$ s.t. $f(x)=x$ and $\mu_x^{g}$ s.t. $g(x)=-x$. Then we have:
\begin{equation*}
\begin{array}{lclcl}
(\traceset, t) \satweak \mu_x^{f}(\epsilon) & \eqdef & \exists x \in [\lb, \ub], x > 0 & \eqdef & [\lb, \ub] \cap (0, +\infty) \neq \emptyset \\
(\traceset, t) \satweak \mu_x^{g}(\epsilon) & \eqdef & \exists x \in [\lb, \ub], x < 0 & \eqdef & [\lb, \ub] \cap (-\infty, 0) \neq \emptyset \\
(\traceset, t) \satstrong \mu_x^{f}(\epsilon) & \eqdef & \forall x \in [\lb, \ub], x > 0 & \eqdef & [\lb, \ub] \subset (0, +\infty) \\
(\traceset, t) \satstrong \mu_x^{g}(\epsilon) & \eqdef & \forall x \in [\lb, \ub], x < 0 & \eqdef & [\lb, \ub] \subset (+\infty, 0) \\
\end{array}
\end{equation*}
$$ (\traceset, t) \satweak \neg \mu_x^{f}(\epsilon) \eqdef \underbrace{\exists x \in [\lb, \ub], x \leq 0}_{[\lb, \ub] \cap (-\infty, 0] \neq \emptyset} \eqdef \underbrace{\neg(\forall x \in [\lb, \ub], x > 0)}_{[\lb, \ub] \nsubset (0, +\infty)}   \eqdef   (\traceset, t) \notsatstrong \mu_x^{f}(\epsilon)$$
\noindent Following the definition of negation for the weak semantics 
we have that:
\begin{equation*}
\begin{array}{lcl}
(\traceset, t) \satweak (\mu_x^{f}(\epsilon) \land \mu_x^{g}(\epsilon)) 
& \eqdef & (\traceset, t) \notsatstrong \neg(\mu_x^{f}(\epsilon) \land \mu_x^{g}(\epsilon))\\
\end{array}
\end{equation*}

\noindent We can show the equivalence consistency using the interval intersection/inclusion interpretation:
$$
(\traceset, t) \satweak (\mu_x^{f}(\epsilon) \land \mu_x^{g}(\epsilon)) 
\eqdef (\traceset, t) \satweak \mu_x^{f}(\epsilon)  \land (\traceset, t) \satweak \mu_x^{g}(\epsilon)
$$
$$
 \underbrace{(\traceset, t) \satweak \mu_x^{f}(\epsilon)}_{[\lb, \ub] \cap (0, +\infty) \neq \emptyset}  \land \underbrace{(\traceset, t) \satweak \mu_x^{g}(\epsilon)}_{[\lb, \ub] \cap (-\infty, 0) \neq \emptyset}
 \eqdef
  \underbrace{(\traceset, t) \notsatstrong \neg \mu_x^{f}(\epsilon)}_{[\lb, \ub] \nsubset (-\infty,0]}  \land \underbrace{(\traceset, t) \notsatstrong \neg \mu_x^{g}(\epsilon)}_{[\lb, \ub] \nsubset [0, +\infty)}
$$

\noindent By applying the De Morgan's laws shown before: 
$$
(\traceset, t) \notsatstrong \neg \mu_x^{f}(\epsilon) \land (\traceset, t) \notsatstrong \neg \mu_x^{g}(\epsilon)
  \eqdef
  (\traceset, t) \notsatstrong \neg(\mu_x^{f}(\epsilon) \land \mu_x^{g}(\epsilon))
$$

}

%\revision{

% \begin{proposition}\label{pro:neg}
% The negation of not strong satisfaction is equivalent to strong satisfaction; the negation of not weak satisfaction is equivalent to weak satisfaction. Formally, we have, 
% \begin{equation*}
% \begin{array}{ll}
% (\traceset, t) \satstrong \neg \neg \varphi &\eqdef  (\traceset, t) \satstrong \varphi \\
% (\traceset, t) \satweak \neg \neg \varphi &\eqdef (\traceset, t) \satweak \varphi \\
% \end{array}
% \end{equation*}
% \end{proposition}
%\begin{proof}
% [Proof of Proposition \ref{pro:neg}]
%Following the negation operation of strong and weak semantics in \defref{def:strongboolean} and \defref{def:weakboolean}, we have, 

%$(\traceset, t) \satstrong \neg \neg \varphi \eqdef (\traceset, t) \notsatweak \neg \varphi \eqdef (\traceset, t) \satstrong \varphi$, and 

%$(\traceset, t) \satweak \neg \neg \varphi \eqdef (\traceset, t) \notsatstrong \neg \varphi \eqdef (\traceset, t) \satweak \varphi$.
%\end{proof}
%}

\begin{proof}
[Proof of \thref{th:strong} and \thref{th:weak}]

% \small
{Mathematically, we want to show that for any $\omega$ and $t$, we have $\forall \conflevel \in \epsilon_s(\varphi, \omega, t), (\omega, t) \satstrong \varphi$ under confidence level $\conflevel$, and for any $\omega$ and $t$, we have $\forall  \conflevel \in \epsilon_w(\varphi, \omega, t), (\omega, t) \satweak \varphi$ under confidence level $ \conflevel$ . We prove the \thref{th:strong} and \thref{th:weak} inductively. Since in the definition of the strong definition and weak definitions refer to each other, we prove them together. We study whether \thref{th:strong} satisfies in every possible case in definition. Then, we finish our proof by the axiom of induction. We omit the cases of $\always$ and $\eventually$ since they can be derived from the case of $\until$.}

\begin{itemize}
% [topsep=0pt, wide=0pt]
% \small
    \item When $\varphi = \mu_\mathsf{x}$, we show $\forall \omega, \forall t, \forall \conflevel \in \epsilon_s(\varphi, \omega, t), (\omega, t) \satstrong \varphi$.
    
   By \defref{def:strongcl}, $\conflevelFun_s(\varphi, \traceset, t) =  
 (0, \int_{\theta_t - \eta}^{\theta_t + \eta} \varphi_t(x) dx), \mathsf{where}\ \eta = \inf\left\{|x - \theta_t| \mathrel{\Big|} f(x) \leq 0 \right\}$. By the definition of confidence interval, we have for $\conflevel \in \conflevelFun_s(\varphi, \traceset, t) $,  $\ub \leq \theta_t + \eta$ and $\lb \geq \theta_t - \eta$, which indicates $ [\lb, \ub] \subseteq [\theta_t - \eta, \theta_t +\eta]$. As $\eta = \inf\left\{|x - \theta_t| \mathrel{\Big|} f(x) \leq 0 \right\}$, we have $\forall x \in [\lb, \ub], f(x) > 0$. Therefore, $(\omega, t) \satstrong \varphi$.
    \item When $\varphi = \mu_\mathsf{x}$, we show $\forall \omega, \forall t, \forall \conflevel \in \epsilon_w(\varphi, \omega, t), (\omega, t) \satweak \varphi$. 
    
    By \defref{def:weakcl}, $\conflevelFun_w(\mu_\mathsf{x}, \traceset, t) = (\int_{\theta_t - \eta}^{\theta_t + \eta} \varphi_t(x) dx, 1),~\mathsf{where}\ \eta = \inf\left\{|x - \theta_t| \mathrel{\Big|} f(x) > 0 \right\}$. By the definition of confidence interval, we have for $\conflevel \in \conflevelFun_s(\varphi, \traceset, t) $,  $\ub \geq \theta_t + \eta$ and $\lb \leq \theta_t - \eta$, which indicates $(\theta_t - \eta, \theta_t +\eta) \subseteq [\lb, \ub]$. Since $\eta = \inf\left\{|x - \theta_t| \mathrel{\Big|} f(x) > 0 \right\}$, we have $\exists x \in [\lb, \ub],  f(x) > 0$. Therefore, $(\omega, t) \satweak \varphi$.
    
    \item When $\varphi = \neg \varphi_1$, we show $\forall \omega, \forall t, \forall \conflevel \in \epsilon_w(\varphi_1, \omega, t), (\omega, t) \satweak \varphi_1 \Rightarrow \forall \omega, \forall t, \forall \conflevel \in \epsilon_s(\varphi, \omega, t)$, $(\omega, t) \satstrong \varphi$.
    
    By \defref{def:strongcl}, we have $\conflevelFun_s(\neg \varphi, \traceset, t) = \conflevelFun_w^c(\varphi_1, \traceset, t)$. Therefore, $\forall \conflevel \in \conflevelFun_s(\neg \varphi, \traceset, t)$, we have $\conflevel \in \conflevelFun_w^c(\varphi_1, \traceset, t)$. Then, by the definition of confidence interval we have $(\omega, t) \notsatweak \varphi_1$ under $\conflevel$. By the  \defref{def:strongboolean}, $(\omega, t) \satstrong \varphi$.

    \item When $\varphi = \neg \varphi_1$, we show $\forall \omega, \forall t, \forall \conflevel \in \epsilon_s(\varphi_1, \omega, t), (\omega, t) \satstrong \varphi_1 \Rightarrow \forall \omega, \forall t, \forall \conflevel \in \epsilon_w(\varphi, \omega, t)$, $(\omega, t) \satstrong \varphi$.
    
    By \defref{def:weakcl}, we have $\conflevelFun_w(\neg \varphi, \traceset, t) = \conflevelFun_s^c(\varphi_1, \traceset, t)$. Therefore, $\forall \conflevel \in \conflevelFun_w(\neg \varphi, \traceset, t)$, we have $\conflevel \in \conflevelFun_s^c(\varphi_1, \traceset, t)$. Then, by the definition of confidence interval we have $(\omega, t) \notsatstrong \varphi_1$ under $\conflevel$. By the definition \ref{def:weakboolean}, $(\omega, t) \satweak \varphi$.
    
    \item $(\varphi = \varphi_1 \land \varphi_2) \land (\forall \omega, \forall t, \forall \conflevel \in \epsilon_s(\varphi_1, \omega, t), (\omega, t) \satstrong \varphi_1) \land (\forall \omega, \forall t, \forall \conflevel \in \epsilon_s(\varphi_2, \omega, t), (\omega, t) \satstrong \varphi_2) \Rightarrow \forall \omega, \forall t, \forall \conflevel \in \epsilon_w(\varphi, \omega, t), (\omega, t) \satstrong \varphi$.
    
    By \defref{def:strongcl}, $\conflevelFun_s(\neg \varphi, \traceset, t) = \conflevelFun_s(\varphi_1, \traceset, t) \cap \conflevelFun_s(\varphi_2, \traceset, t) $. Therefore, $\forall \conflevel \in \conflevelFun_s(\varphi_1 \land \varphi_2, \traceset, t)$, we have $\conflevel \in \conflevelFun_s(\varphi_1, \traceset, t)$ and $\conflevel \in \conflevelFun_s(\varphi_2, \traceset, t)$. Then we have $(\omega, t) \satstrong \varphi_1$ under $\conflevel$ and $(\omega, t) \satstrong \varphi_2$ under $\conflevel$, which indicates $(\omega, t) \satweak \varphi$ by \defref{def:strongboolean}.
    
    \item When $\varphi = \varphi_1 \land \varphi_2$, we show $(\forall \omega, \forall t, \forall \conflevel \in \epsilon_w(\varphi_1, \omega, t), (\omega, t) \satweak \varphi_1) \land (\forall \omega, \forall t, \forall \conflevel \in \epsilon_s(\varphi_2, \omega, t), (\omega, t) \satweak \varphi_2) \Rightarrow \forall \omega, \forall t, \forall \conflevel \in \epsilon_w(\varphi, \omega, t), (\omega, t) \satweak \varphi$.
    
    By \defref{def:weakcl}, $\conflevelFun_w(\neg \varphi, \traceset, t) = \conflevelFun_w(\varphi_1, \traceset, t) \cap \conflevelFun_w(\varphi_2, \traceset, t) $. Therefore, $\forall \conflevel \in \conflevelFun_w(\varphi_1 \land \varphi_2, \traceset, t)$, we have $\conflevel \in \conflevelFun_w(\varphi_1, \traceset, t)$ and $\conflevel \in \conflevelFun_w(\varphi_2, \traceset, t)$. Then we have $(\omega, t) \satweak \varphi_1$ under $\conflevel$ and $(\omega, t) \satweak \varphi_2$ under $\conflevel$, which indicates $(\omega, t) \satweak \varphi$ by \defref{def:weakboolean}.

    \item When $\varphi = \varphi_1 \until \varphi_2$, we show $(\forall \omega, \forall t, \forall \conflevel \in \epsilon_s(\varphi_1, \omega, t), (\omega, t) \satstrong \varphi_1) \land (\forall \omega, \forall t, \forall \conflevel \in \epsilon_s(\varphi_2, \omega, t)$, $(\omega, t) \satstrong \varphi_2) \Rightarrow \forall \omega, \forall t, \forall \conflevel \in \epsilon_s(\varphi, \omega, t), (\omega, t) \satstrong \varphi$.
    
    By \defref{def:strongcl}, we have $\conflevelFun_s( \varphi_1 \until \varphi_2, \traceset, t) =
     \underset{t' \in (t+I)}{\bigcup}
    \left\{\conflevelFun_s(\varphi_2, \traceset, t')\cap(
    \underset{t'' \in (t, t')}{\bigcap}    \conflevelFun_s(\varphi_1, \traceset, t''))\right\} $. Therefore, for $\conflevel \in \epsilon_s(\varphi_1 \until \varphi_2, \traceset, t)$, we have $\exists t' \in (t+I), \conflevel \in \conflevelFun_s(\varphi_2, \traceset, t')\cap(
    \underset{t'' \in (t, t')}{\bigcap}    \conflevelFun_s(\varphi_1, \traceset, t''))$. Therefore, for this $t'$ we have $ \conflevel \in \conflevelFun_s(\varphi_2, \traceset, t')$ and $\forall t'' \in (t, t'), \conflevel \in \conflevelFun_s(\varphi_1, \traceset, t''))$. By the inductive assumption, we have $(\omega, t') \satstrong \varphi_2$ and $\forall t'' \in (t, t'), (\omega, t'') \satstrong \varphi_1$. Finally, by definition \defref{def:strongboolean}, we have $(\omega, t) \satstrong  \varphi_1 \until \varphi_2$.

    \item When $\varphi = \varphi_1 \until \varphi_2$, we show $(\forall \omega, \forall t, \forall \conflevel \in \epsilon_w(\varphi_1, \omega, t), (\omega, t) \satweak \varphi_1) \land (\forall \omega, \forall t, \forall \conflevel \in \epsilon_w(\varphi_2, \omega, t), (\omega, t) \satweak \varphi_2) \Rightarrow \forall \omega, \forall t, \forall \conflevel \in \epsilon_w(\varphi, \omega, t), (\omega, t) \satweak \varphi$.
    
    By \defref{def:weakcl}, we have $\conflevelFun_w( \varphi_1 \until \varphi_2, \traceset, t) =
     \underset{t' \in (t+I)}{\bigcup}
    \left\{\conflevelFun_w(\varphi_2, \traceset, t')\cap(
    \underset{t'' \in (t, t')}{\bigcap}    \conflevelFun_w(\varphi_1, \traceset, t''))\right\} $. Therefore, for $\conflevel \in \epsilon_w(\varphi_1 \until \varphi_2, \traceset, t)$, we have $\exists t' \in (t+I), \conflevel \in \conflevelFun_w(\varphi_2, \traceset, t')\cap(
    \underset{t'' \in (t, t')}{\bigcap}    \conflevelFun_w(\varphi_1, \traceset, t''))$. Therefore, for this $t'$ we have $ \conflevel \in \conflevelFun_w(\varphi_2, \traceset, t')$ and $\forall t'' \in (t, t'), \conflevel \in \conflevelFun_w(\varphi_1, \traceset, t''))$. By the inductive assumption, we have $(\omega, t') \satweak \varphi_2$ and $\forall t'' \in (t, t'), (\omega, t'') \satweak \varphi_1$. Finally, by \defref{def:weakboolean}, we have $(\omega, t) \satweak  \varphi_1 \until \varphi_2$.

\end{itemize}

\end{proof}

% \begin{alignat*}{2}
%     &\conflevelFun_s(\mu_x, \traceset, t) && =  
% [0, \int_{\theta_t - \eta}^{\theta_t + \eta} \varphi_t(x) dx], \mathsf{where}\ \eta = \inf\left\{|x - \theta_t| \mathrel{\Big|} f(x) \leq 0 \right\}
% % \emptyset, \text{if } \eta = 0
%   \\
%     &\conflevelFun_s(\neg \varphi, \traceset, t) && = \conflevelFun_w^c(\varphi, \traceset, t) \\
%     &\conflevelFun_s(\varphi_1 \land \varphi_2, \traceset, t) && =  
%     \conflevelFun_s(\varphi_1, \traceset, t) \cap \conflevelFun_s(\varphi_2, \traceset, t)   \\
%  %   &\conflevelFun_s(\always_I \varphi, \traceset, t) && = \underset{t' \in (t, t+I)}{\cap} 
%  %  \conflevelFun_s(\varphi, \traceset, t')\\
%      &\conflevelFun_s(\always_I \varphi, \traceset, t) && = \underset{t' \in (t+I)}{\cap} 
%   \conflevelFun_s(\varphi, \traceset, t')\\
%       &\conflevelFun_s(\eventually_I \varphi, \traceset, t) && = \underset{t' \in (t+I)}{\cup} 
%   \conflevelFun_s(\varphi, \traceset, t')\\
%     &\conflevelFun_s( \varphi_1 \until \varphi_2, \traceset, t) && =
%     \underset{t' \in (t+I)}{\bigcup}
%     \left\{\conflevelFun_s(\varphi_2, \traceset, t')\cap(
%     \underset{t'' \in (t, t')}{\bigcap} 
%   \conflevelFun_s(\varphi_1, \traceset, t''))\right\} 
%          \\
%  \end{alignat*}

% Intuitively, if there exists a confidence level that guarantees the strong satisfaction, then this requirement can definitely be weakly satisfied

% \small{\textbf{Please ask PC Chairs for the rest of our appendix, which includes the monitoring algorithms for strong and weak satisfaction, and confidence guarantees for strong and weak satisfaction.} } 
\clearpage
% In the appendix, we first present the monitoring algorithms for STL-U strong and weak satisfaction semantics (Algorithm \ref{alg:strongB} and \ref{alg:weakB}).
% % , where $\mathsf{minimize}$ is an local minimization algorithm based on Brent's Method ~\cite{brent2013algorithms}. 
% % \blue{It returns if there exist a point that $f(x)$ equals to 0 with $x\in[\lb,\ub]$. }
% Then we show the algorithms for calculating the confidence levels of strong and weak satisfaction (Algorithm \ref{alg:scf} and \ref{alg:wcf}), respectively. 

%\blue{Then, we give an example to illustrate how to calculate the confidence levels.}
%We use a similar iterative algorithm.

% example for calculating confidence level
% \begin{example}

\noindent
\begin{minipage}[t]{.5\textwidth}
\centering
\begin{algorithm}[H]
\tablefontsize
%   \SetKwFunction{StrongConfidence}{StrongConfidence}
  \SetKwProg{Fn}{Function}{:}{}
  \Fn{$\mathsf{StrongSat}(\varphi,\traceset, t)$}{
      \SetKwInOut{Input}{Input}
      \SetKwInOut{Output}{Output}
      \SetKwFor{Case}{Case}{}{}
      %\Input{STL-U formula $\varphi$, flowpipe $\traceset$, time $t$}
      %\Output{Boolean}
      \Begin{
            \Switch{$\varphi$} {
                  \Case{$\mu_\mathsf{x}(\conflevel)$}{
                
                  \uIf{$ \mathsf{minimize}(f(x), \lb,\ub) > 0$}{ 
                  \Return $\mathsf{True}$ \;
                    }
                  \Else{
                    \Return $\mathsf{False}$ \;
                  }
                  }
                  \Case{$\neg \varphi$}{
                        \Return $\neg \mathsf{WeakSat}$(${\varphi,\traceset, t}$);
                  }
 
                  \Case{$\varphi_1 \land \varphi_2$}{
                        \Return $\mathsf{StrongSat}$(${\varphi,\traceset, t}) \land
                        \mathsf{StrongSat}({\varphi,\traceset, t})$ 
                    }
                    
                    \Case{$\always_I \varphi$}{ 
                    % $\mathsf{Temp} \leftarrow \mathsf{StrongSat}(\varphi, \traceset, t)$
                    
                         \For {$t' \in (t+I)$}
             { 
            %  $\mathsf{Temp} \leftarrow \mathsf{Temp} \land \mathsf{StrongSat}(\varphi, \traceset, t')$ 
            \If{$\neg \mathsf{StrongSat}(\varphi, \traceset, t')$}{
           \Return False; }
              
             }
                         
                         \Return  $\mathsf{True}$ \;
                    }
                    \Case{$\eventually_I \varphi$}{ 
                    % $\mathsf{Temp} \leftarrow \mathsf{StrongSat}(\varphi, \traceset, t)$
                    
                         \For {$t' \in (t+I)$}
             { 
            %  $\mathsf{Temp} \leftarrow \mathsf{Temp} \lor \mathsf{StrongSat}(\varphi, \traceset, t')$
             \If{$\mathsf{StrongSat}(\varphi, \traceset, t')$}{
             \Return True;
             }
              
             }
                         
                         \Return  $\mathsf{False}$ \;
                    }
                    
                    \Case{$\varphi_1 \mathcal{U}_I \varphi_2$}{
                    % $\mathsf{Temp}\leftarrow \mathsf{True}$;
             
             \For {$t'\in (t + I)$}
             {
                  \If {$\mathsf{StrongSat}({\varphi_2,\traceset, t'})$}{
                    %   f := True;
                      
                       \For {$t'' \in [t,t']$}{
                            % $\mathsf{Temp} \leftarrow \mathsf{Temp}  \wedge  \mathsf{StrongSat}({\varphi_1,\traceset, t''})$;
                            
                            \If{$\neg \mathsf{StrongSat}({\varphi_1,\traceset, t''})$}{\Return $\mathsf{False}$ \;}
                       }
                       \Return $\mathsf{True}$ \;
                   }
             }
             \Return $\mathsf{False}$ \;

                    }

          }
      }
    }
\caption{STL-U strong satisfaction monitoring algorithm $\mathsf{StrongSat}({\varphi,\traceset, t})$}
\label{alg:strongB}
\end{algorithm}
\end{minipage}
\hspace{0.2cm}
\begin{minipage}[t]{.5\textwidth}
\centering 
\begin{algorithm}[H]
\tablefontsize
%   \SetKwFunction{StrongConfidence}{StrongConfidence}
  \SetKwProg{Fn}{Function}{:}{}
  \Fn{$\mathsf{WeakSat}(\varphi,\traceset, t)$}{
      \SetKwInOut{Input}{Input}
      \SetKwInOut{Output}{Output}
      \SetKwFor{Case}{Case}{}{}
    %   \Input{STL-U formula $\varphi$, flowpipe $\traceset$, time $t$, Constant $\delta>0$}
      
    %   \Output{Weak Boolean Satisfaction $\mathsf{Boolean}$}
      
      \Begin{
            \Switch{$\varphi$} {
                  \Case{$\mu_\mathsf{x}(\conflevel)$}{
                  \uIf{$ \mathsf{maximize}(f(x), \lb,\ub) > 0$}{
        \Return $\mathsf{True}$ \;
                    }
                  \Else{
                    \Return $\mathsf{False}$ \;
                  }

                  }

                  \Case{$\neg \varphi$}{
                        \Return $\neg \mathsf{StrongSat}$(${\varphi,\traceset, t}$);
                  }
 
                  \Case{$\varphi_1 \land \varphi_2$}{
                        \Return $\mathsf{WeakSat}$(${\varphi,\traceset, t}) \land
                        \mathsf{WeakSat}({\varphi,\traceset, t})$ 
                    }
                    
            %                   \Case{$\always_I \varphi$}{ 
            %         $\mathsf{Temp} \leftarrow \mathsf{WeakSat}(\varphi, \traceset, t)$
                    
            %              \For {$t' \in (t+I)$}
            %  { $\mathsf{Temp} \leftarrow \mathsf{Temp} \land \mathsf{WeakSat}(\varphi, \traceset, t')$ 
              
            %  }
                         
            %              \Return  $\mathsf{Temp}$;
            %         }
            %         \Case{$\eventually_I \varphi$}{ 
            %         $\mathsf{Temp} \leftarrow \mathsf{WeakSat}(\varphi, \traceset, t)$
                    
            %              \For {$t' \in (t+I)$}
            %  { $\mathsf{Temp} \leftarrow \mathsf{Temp} \lor \mathsf{WeakSat}(\varphi, \traceset, t')$ 
              
            %  }
                         
            %              \Return  $\mathsf{Temp}$;
            %         }
                   
            %         \Case{$\varphi_1 \mathcal{U}_I \varphi_2$}{
            %         $\mathsf{Temp}\leftarrow \mathsf{True}$;
             
            %  \For {$t'\in (t + I)$}
            %  {
            %       \If {$\mathsf{WeakSat}({\varphi_2,\traceset, t'})$}{
            %         %   f := True;
                      
            %           \For {$t'' \in [t,t']$}{
            %                 $\mathsf{Temp} \leftarrow \mathsf{Temp}  \wedge  \mathsf{WeakSat}({\varphi_1,\traceset, t''})$;
                            
            %                 \If{$\neg \mathsf{Temp}$}{\textbf{break;}}
            %           }
            %           \If{$\mathsf{Temp}$}{\Return True;}
            %       }
            %  }
            %  \Return False;

            %         }

                   \Case{$\always_I \varphi$}{ 
                    % $\mathsf{Temp} \leftarrow \mathsf{StrongSat}(\varphi, \traceset, t)$
                    
                         \For {$t' \in (t+I)$}
             { 
            \If{$\neg \mathsf{WeakSat}(\varphi, \traceset, t')$}{
           \Return False; }
              
             }
                         
                         \Return  $\mathsf{True}$ \;
                    }
                    \Case{$\eventually_I \varphi$}{ 
                    
                         \For {$t' \in (t+I)$}
             { 
             \If{$\mathsf{WeakSat}(\varphi, \traceset, t')$}{
             \Return True;
             }
              
             }
                         
                         \Return  $\mathsf{False}$ \;
                    }
                    
                    \Case{$\varphi_1 \mathcal{U}_I \varphi_2$}{
                    % $\mathsf{Temp}\leftarrow \mathsf{True}$;
             
             \For {$t'\in (t + I)$}
             {
                  \If {$\mathsf{WeakSat}({\varphi_2,\traceset, t'})$}{
                    %   f := True;
                      
                       \For {$t'' \in [t,t']$}{
                            
                            \If{$\neg \mathsf{WeakSat}({\varphi_1,\traceset, t''})$}{\Return $\mathsf{False}$ \;}
                       }
                       \Return $\mathsf{True}$ \;
                   }
             }
             \Return $\mathsf{False}$ \;

                    }

          }
      }
    }
\caption{STL-U weak satisfaction monitoring algorithm $\mathsf{WeakSat}({\varphi,\traceset, t})$}
\label{alg:weakB}
\end{algorithm}
\end{minipage}

\noindent
\begin{minipage}[t]{.5\textwidth}
\centering
\begin{algorithm}[H]
\tablefontsize
  \SetKwProg{Fn}{Function}{:}{}
  \Fn{$\mathsf{StrongConfidenceLevel}({\varphi,\traceset, t})$}{
      \SetKwInOut{Input}{Input}
      \SetKwInOut{Output}{Output}
      \SetKwFor{Case}{Case}{}{}
    %   \Input{STL-U Requirement $\varphi$, flowpipe $\traceset = \{\Phi_t\}$, time $t$}
      
    %   \Output{Confidence Level of Strong Satisfaction $\mathsf{Float Interval:} \conflevelFun_s \subseteq [0,1] $}
      
      \Begin{
            \Switch{$\varphi$} {
                  \Case{$\mu_\mathsf{x}$}{
        %         %   $\rho^+ \leftarrow \Phi_\conflevel^+(t)$ \\
        %           \uIf{$\theta_t > 0$}{
        %           $\conflevel \leftarrow \int_0^{2\theta} \Phi_t(x) dx$\\
        %           $\conflevelFun_s \leftarrow [0, \conflevel]$\\
        % \Return  $\conflevelFun_s$\;
        %             }
                    
        %           \Else{
        %             \Return $\emptyset$ \;
        %           }
        $\eta \leftarrow \inf\left\{|x - \theta_t| \mathrel{\Big|} f(x) \leq 0 \right\}$  \\
        \Return $(0, \int_{\theta_t - \eta}^{\theta_t + \eta} \Phi_t(x) dx)$; 
        
                  }

                  \Case{$\neg \varphi$}{
                  $\conflevelFun_s \leftarrow \mathsf{WeakConfidenceLevel}({\varphi,\traceset, t})^C$\\
                        \Return $\mathsf{\conflevelFun_s}$;
                  }
 
                  \Case{$\varphi_1 \land \varphi_2$}{
                        \Return $\mathsf{StrongConfidenceLevel}$(${\varphi_1,\traceset, t}) \cap \mathsf{StrongConfidenceLevel}({\varphi_2,\traceset, t})$ 
                    }
                    
                    \Case{$\always_I \varphi$}{ 
                    $\conflevelFun_s = \mathsf{StrongConfidenceLevel}(\varphi, \Omega, 0)$
                    
                         \For {$t' \in (t+I)$}
             { $\conflevelFun_s \leftarrow \conflevelFun_s \cap \mathsf{StrongConfidenceLevel}(\varphi, \Omega, t')$
              
             }
                         
                         \Return  $\conflevelFun_s$;
                    }
                
                \Case{$\eventually_I \varphi$}{ 
                    $\conflevelFun_s \leftarrow \mathsf{StrongConfidenceLevel}(\varphi, \Omega, 0)$
                    
                         \For {$t' \in (t+I)$}
             { $\conflevelFun_s \leftarrow \conflevelFun_s \cup \mathsf{StrongConfidenceLevel}(\varphi, \Omega, t')$
              
             }
                         
                         \Return  $\conflevelFun_s$;
                    }
        
          \Case{$\varphi_1 \mathcal{U}_I \varphi_2$}{
                    % $\mathsf{StrongSat}\leftarrow \mathsf{True}$;
             $\conflevelFun_s  \leftarrow \emptyset$\\
             \For {$t'\in (t + I)$}
             {
            $\conflevelFun_s'  \leftarrow \mathsf{StrongConfidenceLevel}({\varphi_2,\traceset, t'})$\\
            \For {$t'' \in [t,t']$}{
                $\conflevelFun_s'  \leftarrow \conflevelFun_s'\cap \mathsf{StrongConfidenceLevel}({\varphi_1,\traceset, t''})$             
                       }
            $\conflevelFun_s  \leftarrow \conflevelFun_s \cup \conflevelFun_s' $            
                   
             }
        
            \Return  $\conflevelFun_s$;

                    }

          }
      }
    }
\caption{Confidence Level of Strong Satisfaction $\mathsf{StrongConfidenceLevel}({\varphi,\traceset, t})$}
\label{alg:scf}
\end{algorithm}
\end{minipage}
\hspace{0.2cm}
\begin{minipage}[t]{.5\textwidth}
\centering
\begin{algorithm} [H]
\tablefontsize
  \SetKwProg{Fn}{Function}{:}{}
  \Fn{$\mathsf{WeakConfidenceLevel}({\varphi,\traceset, t})$}{
      \SetKwInOut{Input}{Input}
      \SetKwInOut{Output}{Output}
      \SetKwFor{Case}{Case}{}{}
    %   \Input{STL-U Requirement $\varphi$, flowpipe $\traceset = \{\Phi_t\}$, time $t$}
      
    %   \Output{Confidence Level of Weak Satisfaction $\mathsf{Float Interval:} \conflevelFun_w \subseteq [0,1] $}
      
      \Begin{
            \Switch{$\varphi$} {
                  \Case{$\mu_\mathsf{x}$}{
              
        %           \uIf{$\theta_t < 0$}{
        %           $\conflevel \leftarrow \int_0^{2\theta} \Phi_t(x) dx$\\
        %           $\conflevelFun_w \leftarrow [\conflevel, 1)$\\
        % \Return  $\conflevelFun_w$\;
        %             }
                    
        %           \Else{
        %             \Return $[0,1]$ \;
        %           }
         $\eta \leftarrow \inf\left\{|x - \theta_t| \mathrel{\Big|} f(x) > 0 \right\}$  \\
        \Return $(\int_{\theta_t - \eta}^{\theta_t + \eta} \Phi_t(x) dx, 1)$ \; 
       
                  }

                  \Case{$\neg \varphi$}{
                  $\conflevelFun_w \leftarrow \mathsf{StrongConfidenceLevel}({\varphi,\traceset, t})^C$\\
                        \Return $\mathsf{\conflevelFun_w}$;
                  }
 
                  \Case{$\varphi_1 \land \varphi_2$}{
                        \Return $\mathsf{WeakConfidenceLevel}$(${\varphi_1,\traceset, t}) \cap \mathsf{WeakConfidenceLevel}({\varphi_2,\traceset, t})$ 
                    }
                    
                    \Case{$\always_I \varphi$}{ 
                    $\conflevelFun_w \leftarrow \mathsf{WeakConfidenceLevel}(\varphi, \Omega, 0)$
                    
                         \For {$t' \in (t+I)$}
             { $\conflevelFun_w \leftarrow \conflevelFun_w \cap \mathsf{WeakConfidenceLevel}(\varphi, \Omega, t')$
              
             }
                         
                         \Return  $\conflevelFun_w$;
                    }
                    
                    \Case{$\eventually_I \varphi$}{ 
                    $\conflevelFun_w \leftarrow \mathsf{WeakConfidenceLevel}(\varphi, \Omega, 0)$
                    
                         \For {$t' \in (t+I)$}
             { $\conflevelFun_w \leftarrow \conflevelFun_w \cup \mathsf{WeakConfidenceLevel}(\varphi, \Omega, t')$
              
             }
                         
                         \Return  $\conflevelFun_w$;
                    }
        
          \Case{$\varphi_1 \mathcal{U}_I \varphi_2$}{
                    % $\mathsf{StrongSat}\leftarrow \mathsf{True}$;
             $\conflevelFun_w  \leftarrow \emptyset$\\
             \For {$t'\in (t + I)$}
             {
            $\conflevelFun_w'  \leftarrow \mathsf{WeakConfidenceLevel}({\varphi_2,\traceset, t'})$\\
            \For {$t'' \in [t,t']$}{
                $\conflevelFun_w'  \leftarrow \conflevelFun_w'\cap \mathsf{WeakConfidenceLevel}({\varphi_1,\traceset, t''})$             
                       }
            $\conflevelFun_w  \leftarrow \conflevelFun_w \cup \conflevelFun_w' $            
                   
             }
        
            \Return  $\conflevelFun_w $;

                    }

          }
      }
    }
\caption{Confidence Level of Weak Satisfaction $\mathsf{WeakConfidenceLevel}({\varphi,\traceset, t})$}
\label{alg:wcf}
\end{algorithm}
\end{minipage}

\end{document}